\documentclass[runningheads]{llncs}

\usepackage{eccv}

\usepackage{eccvabbrv}

\usepackage{graphicx}
\usepackage{booktabs}
\usepackage{multirow}
\usepackage{booktabs}
\usepackage{array}
\usepackage{tabularx}
\usepackage{makecell}
\usepackage{subcaption}
\usepackage{adjustbox}
\usepackage{float}
\usepackage{wrapfig}
\usepackage{enumitem}
\usepackage{bm}
\usepackage{amssymb}
\usepackage{circledsteps}

\usepackage[accsupp]{axessibility}  

\usepackage{hyperref}

\usepackage{orcidlink}
\usepackage{marvosym}
\usepackage[table]{xcolor}
\definecolor{gray94}{gray}{.94}
\definecolor{bestblue}{RGB}{184, 215, 238}
\definecolor{secondblue}{RGB}{226, 239, 249}
\newcommand{\bestcell}[1]{\cellcolor{bestblue}#1}
\newcommand{\secondcell}[1]{\cellcolor{secondblue}#1}
\newcommand{\boldparagraph}[1]{%
  \paragraph{\textbf{#1}}%
}

\begin{document}

\title{\textsc{ChronoFlow-Policy} \\
Unifying Past-Current-Future Interaction Flow \\in Visuomotor Policy Learning } 

\titlerunning{ChronoFlow-Policy}



\author{
Bokai Lin\inst{1,2}$^{*}$\orcidlink{0009-0008-5789-3563} \and
Yifu Xu\inst{1}$^{*}$\orcidlink{0009-0000-2122-4386}\and
Xinyu Zhan\inst{1}\orcidlink{0009-0004-7859-2592}\and
Hongjie Fang\inst{1}\orcidlink{0000-0002-6309-1160}\and
Jialin Tian\inst{1}\orcidlink{0009-0006-5527-3091} \and
Fu-Cheng Zhang\inst{2}\orcidlink{0009-0007-2610-902X} \and
Yong-Lu Li\inst{1,2}\orcidlink{0000-0003-0478-0692} \and
Cewu Lu\inst{1,2,3}\orcidlink{0000-0003-1533-8576} \and
Lixin Yang\inst{1}\textsuperscript{\Letter}\orcidlink{0000-0001-6366-3192}
}

\authorrunning{B. Lin et al.}

\institute{
Shanghai Jiao Tong University, China
\and
Shanghai Innovation Institute, China
\and
Noematrix, China \\
\email{19821172068@sjtu.edu.cn, siriusyang@sjtu.edu.cn}
}




\maketitle

\begingroup
\renewcommand{\thefootnote}{}
\footnotetext{\hspace{-1.8em}$^{*}$Equal contribution. \textsuperscript{\Letter}Corresponding author.}
\endgroup

\begin{center}
\includegraphics[width=\textwidth]{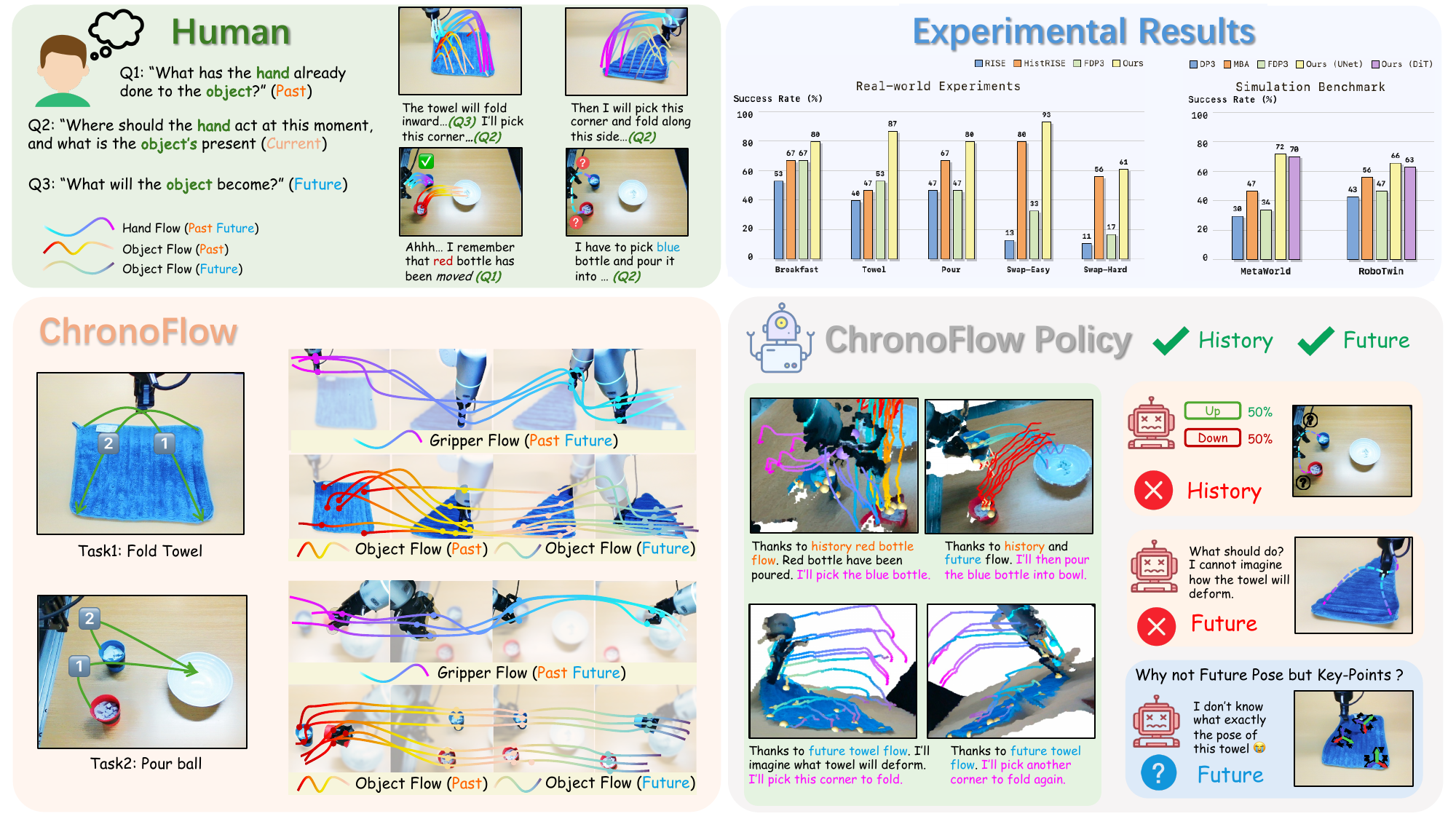}
\captionof{figure}{
 \textbf{Motivation and Overview of ChronoFlow.}
  \textbf{\textcolor[RGB]{76, 123, 49}{Human}} intuition solves manipulation tasks by reasoning over \emph{past--current--future} interaction: recalling how the hand and object have moved, perceiving their current relation, and anticipating how the object should move next.
  Inspired by this temporal reasoning, \textbf{\textcolor[RGB]{231, 173, 138}{ChronoFlow}} provides an intermediate representation for robot that encodes unified object-gripper keypoint flows across time.
  \textbf{\textcolor[RGB]{174, 174, 174}{ChronoFlow-Policy}} leverages ChronoFlow as a co-training objective to learn actions, improving performance on both simulation and real-world manipulation.
  }
\label{fig:teaser}
\end{center}

\begin{abstract}
  Visual signals play a crucial role in policy learning by enabling models to capture object motion and interaction dynamics.
  Just as humans reason about actions using both past experience and anticipated outcomes, effective policies should integrate past interactions with future predictions.
  However, existing visuomotor policies typically model either historical context or future dynamics in isolation,  lacking a unified temporal representation of interaction dynamics.
  In this work, we introduce \textbf{ChronoFlow}, a temporally unified representation that captures \textbf{past, current, and future} interaction dynamics through sparse 3D keypoints of both objects and the gripper.
  Based on this representation, we propose \textbf{ChronoFlow-Policy}, a diffusion-based visuomotor policy that jointly learns ChronoFlow and action sequences through a co-training objective.
  Experiments on 14 simulated tasks and 5 real-world manipulation tasks demonstrate that ChronoFlow-Policy consistently outperforms strong diffusion-policy baselines and improves robustness in long-horizon and non-Markovian manipulation scenarios.
  Code and models will be released at \url{https://github.com/The-kamisato-Sii/ChronoFlow-Policy}.
\end{abstract}

\section{Introduction}
\label{sec:intro}

Recent advances in world action models have begun to couple action generation with predictive visual modeling.
For example, UWM~\cite{zhu2025uwm} and DreamZero~\cite{dreamzero2025} use structured visual prediction as auxiliary supervision, indicating that action generation can be improved by temporally aligned visual targets.

A complementary question is what form of visual supervision is better suited for embodied manipulation. Compared with image-space prediction alone, depth-aware 3D representations lift visual supervision into physical space. For example, object poses and scene point flows can describe how task-relevant entities move and interact over time.
In visuomotor policy learning, existing efforts along this direction mainly follow two routes.
One route predicts \textbf{future state evolution}, using object poses~\cite{su2025motionactiondiffusingobject, hsu2025spotse3posetrajectory} or scene-level point cloud dynamics~\cite{noh20253dflowdiffusionpolicy,huang2026pointworldscaling3dworld} to provide foresight for action generation.
The other route incorporates \textbf{historical context}, motivated by the non-Markovian nature of many manipulation tasks~\cite{diffusion_forcing, ptp, chen2025historyawarevisuomotorpolicylearning}, and uses past observations for long-horizon reasoning~\cite{memoryvla, chen2025historyawarevisuomotorpolicylearning,zheng2025tracevlavisualtraceprompting} . Despite their complementary roles, future prediction and history modeling are typically studied separately.

In this work, we propose \textbf{ChronoFlow}, a temporally unified flow representation that spans \textbf{past}, \textbf{current} and \textbf{future}.
ChronoFlow represents \textbf{both objects and the gripper} (interaction-centric) as sparse 3D keypoints evolving over time, capturing their coupled motion in a shared spatial frame.
ChronoFlow offers two practical advantages:
\begin{enumerate}[label=\textbf{\arabic*.}]
    \item it is supported by recent advances in 3D point tracking~\cite{xiao2025spatialtrackerv23dpointtracking, zhang2025tapip3dtrackingpointpersistent} (\eg D4RT~\cite{zhang2026d4rt}), which make it feasible to extract reliable historical traces from RGBD videos;
    \item it focuses on task-relevant object-gripper keypoints, providing a compact representation that filters redundant scene motion while preserving interaction structure.
\end{enumerate}

Based on ChronoFlow, we propose \textbf{ChronoFlow-Policy}, which couples action generation with ChronoFlow prediction in a unified diffusion framework. By requiring the diffusion backbone to recover interaction-flow trajectories, the auxiliary objective encourages the shared latent representation to encode temporally grounded manipulation dynamics rather than action labels alone. The policy then decodes actions from this shared representation using a lightweight transformer-based decoder.

We evaluate ChronoFlow-Policy on 14 simulated tasks from MetaWorld~\cite{yu2021metaworldbenchmarkevaluationmultitask} and RoboTwin 2.0~\cite{chen2025robotwin20scalabledata}, together with 5 real-world manipulation tasks.
Our method is compared against representative visuomotor policies, including strong 3D imitation policies like
DP3~\cite{ze20243ddiffusionpolicygeneralizable} and RISE~\cite{wang2024rise}.
We further compare with history-aware approaches such as HistRISE~\cite{chen2025historyawarevisuomotorpolicylearning},
as well as policies that incorporate future dynamics modeling, such as MBA~\cite{su2025motionactiondiffusingobject} and 3D-FDP~\cite{noh20253dflowdiffusionpolicy}.

ChronoFlow-Policy achieves consistent performance improvements across both simulation and real-world tasks.
In particular, in the real-world tasks, \textit{Swap-Easy} and \textit{Swap-Hard} require the policy to remember earlier object state when selecting later actions; historical ChronoFlow traces provide this memory and improve stage-dependent decision making.
\textit{Fold Towel} further shows that ChronoFlow can model deformable-object dynamics through interaction flows.
On the long-horizon task \textit{Prepare Breakfast}, ChronoFlow-Policy maintains strong performance across sequential stages, indicating that interaction-flow prediction helps preserve task progress during multi-step execution.

Our contribution are threefolds:
\begin{itemize}
    \item We propose ChronoFlow, a compact interaction-centric 3D keypoint representation that unifies past-current-future dynamics.
    \item We propose ChronoFlow-Policy, a diffusion-based visuomotor policy that couples ChronoFlow prediction with action generation through a co-training objective, enabling actions to be decoded from a shared latent representation.
    \item We design a suite of long-horizon, non-Markovian, and non-rigid real-world manipulation tasks to assess temporal interaction modeling under stage-dependent decisions, deformable-object dynamics, and multi-step execution.
    ChronoFlow-Policy shows consistent gains across these settings, demonstrating the practical value of unified past-current-future interaction modeling.
\end{itemize}

\section{Related work}

\subsubsection{Visuomotor Policy with Future Integration.}
Recent work in visuomotor policy learning increasingly incorporates future representations --- which estimate how the scene may evolve --- as auxiliary supervisions to improve action prediction. Some approaches predict future observations directly in the visual domain, where policies are conditioned on predicted frames or latent video representations.
Such methods employ video diffusion models~\cite{hu2025videopredictionpolicygeneralist, pai2025mimicvideovideoactionmodelsgeneralizable} or jointly model latent video features with actions in a unified sequence~\cite{dreamzero2025, lingbot-va2026}.
While effective at capturing long-horizon temporal patterns, these representations remain entangled with appearance information and often lack explicit spatial structure, which can introduce redundant signals unrelated to physical interaction.

Another line of work models future dynamics using more structured and abstract representations.
Some methods predict object pose trajectories~\cite{su2025motionactiondiffusingobject} or condition policies on sequences of object poses~\cite{hsu2025spotse3posetrajectory}, enabling reasoning over object motion in SE(3).
However, pose-based representations are less suitable for deformable objects and complex multi-object interactions.
Other works represent future dynamics using object-level motion fields, such as object flow~\cite{bharadhwaj2024track2actpredictingpointtracks, wen2024anypointtrajectorymodelingpolicy, xu2024flowcrossdomainmanipulationinterface, eisner2024flowbot3dlearning3darticulation, huang2026pointworldscaling3dworld, su2025freqpolicyefficientflowbasedvisuomotor}.
Some approaches also model gripper flow to capture end-effector motion~\cite{gkanatsios20253dflowmatchactorunified}.
However, these methods typically model object and gripper motion separately, without explicitly capturing their interaction dynamics.
More recently, several methods adopt scene flow to estimate dense point trajectories of the entire scene~\cite{noh20253dflowdiffusionpolicy, huang2026pointworldscaling3dworld,wang2026lamp}, using optical flow~\cite{huang2022flowformertransformerarchitectureoptical, shi2023flowformermaskedcostvolume} or point tracking~\cite{xiao2025spatialtrackerv23dpointtracking, zhang2025tapip3dtrackingpointpersistent}.
While expressive, such dense representations may include redundant motion signals unrelated to task-relevant interactions.

In contrast, we introduce an interaction-centric keypoint flow representation that jointly models past and future trajectories of both objects and grippers. 
By focusing on interaction-relevant keypoints, our formulation avoids redundant information while explicitly capturing gripper-object dynamics, enabling temporally consistent reasoning for complex manipulation.

\subsubsection{Visuomotor Policy with History Integration.}
Prior work explores different ways to encode historical observations for visuomotor policies. 
Video-based representations capture long temporal horizons but incur high computational cost and redundancy~\cite{li2025cronusvlaefficientrobustmanipulation,hu2025videopredictionpolicygeneralist}, while keyframe-based or compressed representations improve efficiency at the risk of missing subtle temporal transitions~\cite{fang2025sam2actintegratingvisualfoundation,tang2025adaptivekeyframesamplinglong,yu2025framevoyagerlearningqueryframes}. 
Hybrid designs combine multiple abstraction levels, yet often preserve visually salient but task-irrelevant information~\cite{jin2024videolavitunifiedvideolanguagepretraining,SlowFast-LLaVA-1.5,shi2026memoryvlaperceptualcognitivememoryvisionlanguageaction,torne2026memmultiscaleembodiedmemory}. 
More recently, several methods like TraceVLA~\cite{zheng2025tracevlavisualtraceprompting} and HistRISE~\cite{chen2025historyawarevisuomotorpolicylearning} inject historical information through object-centric representations, for example by tracking object points over time and using them as historical context for policy learning. 
Most existing history representations focus on encoding past observations as contextual inputs for policy learning.
However, these representations typically do not explicitly model object-gripper interactions, which play a central role in many manipulation tasks.

On the contrary, our method explicitly extracts interaction-centric keypoints that capture object-gripper contact and motion patterns.
Moreover, beyond encoding history, we also predict future interaction trajectories and use them as a co-training objective.
This design provides structured supervision on how past interactions evolve into future states, encouraging the model to learn unified past-current-future interaction dynamics for manipulation.

\section{Methodology}

\subsection{Problem Setup}
We study visuomotor policy learning for robot manipulation from demonstrations.
At each timestep $t$, the robot receives an observation
$\bm{o}_t = (\bm{p}_t, \bm{q}_t)$.
Here $\bm{p}_t \in \mathbb{R}^{N \times 6}$ denotes a scene-level point cloud captured from a single RGB-D camera, where each point is represented by its 3D coordinates and color $(x,y,z,r,g,b)$.
The number of points $N$ may vary across timesteps.
When available, $\bm{q}_t \in \mathbb{R}^{d_q}$ denotes the robot proprioceptive state, such as joint positions or end-effector pose.
Given the current observation, the policy predicts a sequence of future low-level actions over a fixed horizon $H$, $\bm{o}_t \mapsto \bm{a}_{t:t+H}.$ where each $\bm{a}_\tau \in \mathbb{R}^{d_a}$ corresponds to continuous control commands, such as end-effector pose increments.

Learning such a policy from demonstrations is challenging in complex manipulation scenarios. Observations from a single timestep provide only partial information about the interaction state, while many manipulation tasks exhibit strong temporal dependencies due to contact dynamics and staged object interactions.
As a result, policies trained solely on instantaneous observations often struggle to capture the underlying spatiotemporal structure of object–robot interactions.
In this work, we address this challenge by introducing an interaction-centric representation that explicitly models the temporal evolution of task-relevant motions.

\subsection{ChronoFlow: Past-Current-Future Interaction Flow}
\label{subsec: chronoflow}
To explicitly model interaction dynamics in manipulation tasks, we introduce \textbf{ChronoFlow}, a unified representation of past-current-future interaction based on sparse 3D keypoint trajectories.

At each timestep $t$, ChronoFlow represents the interaction state using a set of keypoints defined on both the robot gripper and task-relevant objects.
Specifically, we denote the gripper keypoints as $\bm{P}^{g}_t \in \mathbb{R}^{N_g \times 3}$ and the object keypoints as $\bm{P}^{o}_t \in \mathbb{R}^{N_o \times 3}$, where $N_g$ and $N_o$ are the numbers of gripper and object keypoints, respectively.
Their union $\bm{P}_t = \bm{P}^{g}_t \cup \bm{P}^{o}_t$ provides a sparse abstraction of the physical interaction state at time $t$.

\boldparagraph{Gripper Keypoints Flow.}
For the robot gripper, we predefine $N_g$ canonical keypoints on the gripper geometry.
These keypoints are rigidly attached to the tool center point (TCP).
Given the TCP pose at each timestep, we compute the 3D positions of all gripper keypoints via rigid transformation, producing a temporally consistent gripper trajectory in the world coordinate frame.

\boldparagraph{Object Keypoints Flow.}
For manipulated objects, we first segment task-relevant objects from the initial observation using 3D segmentation SAM-2~\cite{ravi2024sam2segmentimages}.
For each segmented object, we apply Farthest Point Sampling (FPS) to obtain a dense set of candidate keypoints.
Aggregating across objects yields a pool of object keypoints that sparsely represent the scene geometry.
These keypoints are tracked over time using TAPIP3D~\cite{zhang2025tapip3dtrackingpointpersistent}, producing temporally consistent object-centric trajectories.

The resulting ChronoFlow trajectories provide a compact representation of interaction dynamics.
Since the gripper flow is structurally aligned with robot actions, perturbing ChronoFlow implicitly injects noise into both the future object motion and action.
This interaction-centric formulation naturally leads to joint flow matching.
In the next section, we describe how these trajectories are incorporated into a joint flow-matching visuomotor policy.

\begin{figure*}[t]
  \centering
  \includegraphics[width=\textwidth]{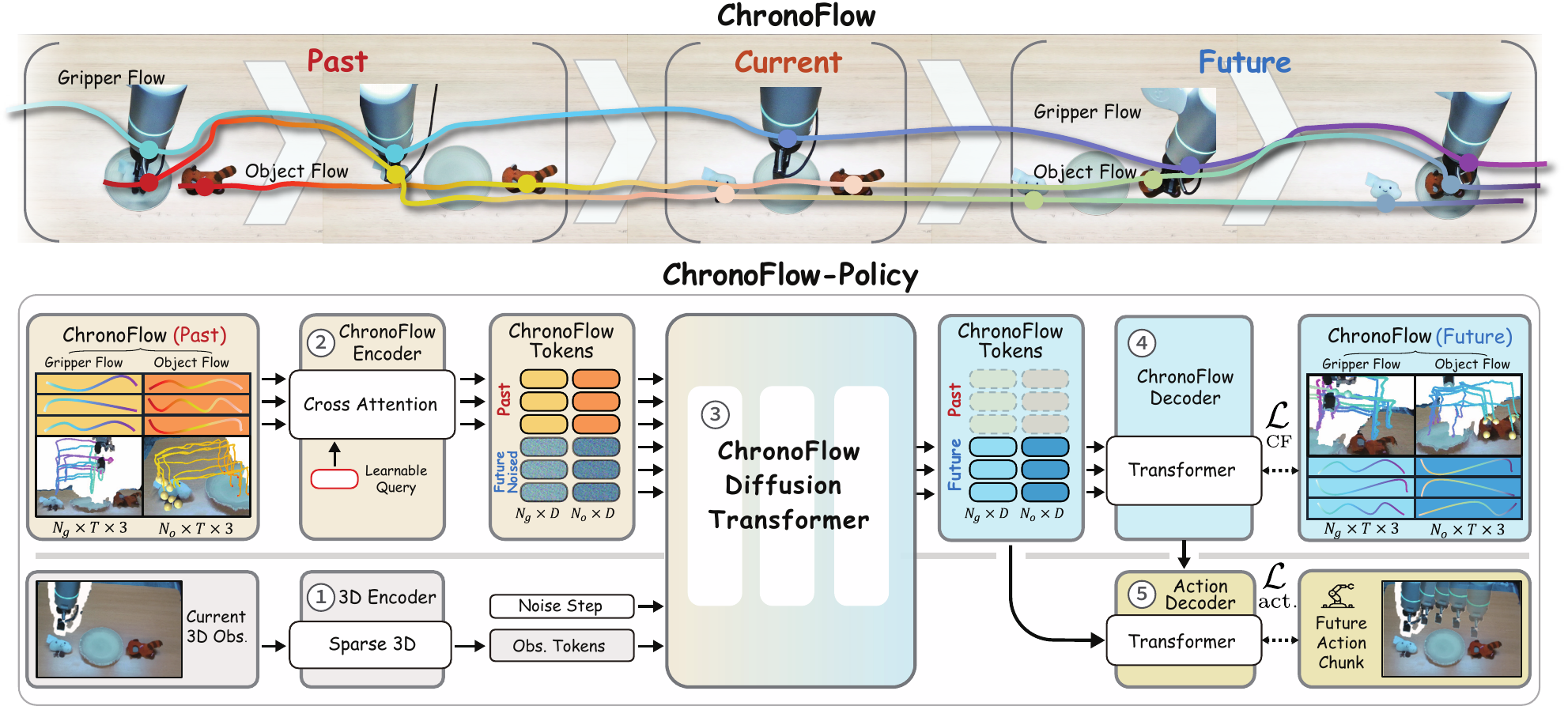}
  \caption{
  Overview of \textbf{ChronoFlow-Policy}.
    \textbf{Top:} Illustration of ChronoFlow: unified past-current-future object-gripper keypoint flows.
    \textbf{Bottom:} The network architecture.
  }
  \label{fig:method_overview}
\end{figure*}

\subsection{ChronoFlow-Policy}

Based on the ChronoFlow representation, we propose \textbf{ChronoFlow-Policy}, a diffusion-based visuomotor policy that explicitly models interaction dynamics and leverages them for action generation.

\boldparagraph{Overview.}
Instead of directly predicting actions from current visual observations, 
ChronoFlow-Policy first models the future evolution of interaction-centric keypoints and then generates robot actions conditioned on the predicted interaction trajectory.
Formally, we factorize the policy as
\begin{equation}
\pi(\bm{a}_{t:t+H}\mid \bm{o}_t,\bm{P}_{:t})
=
\pi_\phi(\bm{a}_{t:t+H}\mid \bm{P}_{t:t+H})
\,
\pi_\theta(\bm{P}_{t:t+H} \mid \bm{o}_t,\bm{P}_{:t})
\label{eq:factorization}
\end{equation}
where $\bm{P}_{t:t+H}$ denotes the ChronoFlow trajectory describing the future motion, and $\bm{P}_{:t} = \{\bm{P}_\tau\}_{\tau < t}$ denotes the historical ChronoFlow trajectories constructed from past observations.
The first term $\pi_\theta$ in Eq.\ref{eq:factorization} models the evolution of interaction dynamics, while the second term  $\pi_\phi$ maps the predicted interaction trajectory to robot actions. 

\boldparagraph{Diffusion-based Interaction Modeling.}
To model the distribution
$\pi_\theta(\bm{P}_{t:t+H}\mid\bm{o}_t,\bm{P}_{:t})$,
we employ a conditional diffusion process over ChronoFlow trajectories.
Starting from Gaussian noise, the model iteratively refines a noisy trajectory through a sequence of denoising steps:
\begin{equation}
\bm{P}^{k-1}_{t:t+H}
=
\alpha_k \bm{P}^{k}_{t:t+H}
-
\gamma_k \epsilon_\theta(\bm{P}^{k}_{t:t+H},\,\bm{o}_t,\,\bm{P}_{:t},\,k)
+
\sigma_k \mathcal{N}(0,I),
\label{eq:flow_matching}
\end{equation}
where $\epsilon_\theta$ denotes the denoising network, $k \in (0,1)$ denotes the diffusion timestep, and 
$\{\alpha_k,\gamma_k,\sigma_k\}$ define the diffusion noise schedule.

\boldparagraph{\textup{\Circled{1}} Observation Encoder.}
Observation encoder maps the current 3D point cloud observation to a condition feature $\bm{c}_t=E_{\mathrm{obs}}(\bm{o}_t)$.
We use a PointNet-style point-cloud encoder in simulation and a sparse 3D encoder in real-world experiments.
When proprioception is available, an MLP embeds it and fuses it with the point-cloud feature.

\boldparagraph{\textup{\Circled{2}} ChronoFlow Encoder.}
ChronoFlow encoder converts historical ChronoFlow and the noisy future ChronoFlow into \textbf{ChronoFlow tokens}.
Given the future ChronoFlow trajectory
$\bm{P}_{t:t+H}\in\mathbb{R}^{(N_g+N_o)\times H\times 3}$,
we perturb it with noise at flow-matching step $k$, obtaining $\bm{P}^{k}_{t:t+H}$.
The multi-head cross-attention module (MCA in \cref{eq:chrono_flow_encoder}) uses learnable interaction queries
$\bm{Q} \in \mathbb{R}^{(N_g+N_o)\times D}$
to jointly encode the historical ChronoFlow $\bm{P}_{:t}$ and the noisy future ChronoFlow $\bm{P}^{k}_{t:t+H}$
into ChronoFlow tokens $\bm{Z}^{k}_{t} \in \mathbb{R}^{(N_g+N_o)\times D}$:
\begin{equation}
    \bm{Z}^{k}_{t}
    =
    \mathrm{MCA}
    \left(
    \bm{Q},
    \left[
    \bm{P}_{:t};
    \bm{P}^{k}_{t:t+H}
    \right] \cdot
    \bm{W}_{K}, 
    \left[
    \bm{P}_{:t};
    \bm{P}^{k}_{t:t+H}
    \right] \cdot
    \bm{W}_{V}
    \right),
\label{eq:chrono_flow_encoder}
\end{equation}
where $\bm{W}_{K}$ and $\bm{W}_{V}$ are the key and value projection matrices, respectively.
Each compact ChronoFlow token corresponds to one gripper or object keypoint.
$\bm{Z}^{k}_{t}$ serves as the compact latent representation used by the flow-matching model $\epsilon_{\theta} \left(\bm{P}^{k}_{t:t+H}, \bm{o}_t, \bm{P}_{:t}, k\right)$ in Eq.~\ref{eq:flow_matching}.

\boldparagraph{\textup{\Circled{3}} Joint Diffusion Denoising.}
The backbone of ChronoFlow-Policy is a diffusion module, implemented either as a Unet or a Diffusion Transformer (DiT).
Given the observation tokens, the noise timestep, and the ChronoFlow tokens $\bm{Z}^{k}_{t}$, the model performs conditional denoising in the ChronoFlow token space. This process progressively refines $\bm{Z}^{k}_{t}$ into clean interaction tokens $\bm{Z}^{0}_{t}$, which serve as the shared latent representation for subsequent ChronoFlow prediction and action decoding.

\boldparagraph{\textup{\Circled{4}} ChronoFlow Decoder.}
After denoising step, the ChronoFlow decoder converts the denoised tokens $\bm{Z}^{0}_{t}$ into future ChronoFlow using a lightweight Transformer decoder:
\begin{equation}
    \hat{\bm{P}}_{t:t+H}
    =
    D_{\mathrm{CF}}(\bm{Z}^{0}_{t})
    \in
    \mathbb{R}^{(N_g+N_o)\times (H\cdot 3)} .
\end{equation}
The output is reshaped into
$\hat{\bm{P}}_{t:t+H}\in\mathbb{R}^{(N_g+N_o)\times H\times 3}$,
representing future 3D trajectories of gripper and object keypoints.

\boldparagraph{\textup{\Circled{5}} Action Decoder.}
Action decoder predicts robot actions from the decoded ChronoFlow trajectory and the diffusion hidden state.
We concatenate $\hat{\bm{P}}_{t:t+H}$ with the clean ChronoFlow tokens $\bm{Z}_{t}^{0}$ and feed them into a lightweight transformer decoder $\mathcal{A}_\phi(\cdot)$ for action prediction: $\hat{\bm{a}}_{t:t+H} = \mathcal{A}_\phi([\bm{Z}_{t}^0; \hat{\bm{P}}_{t:t+H}]).$

\boldparagraph{Training Objectives.}
The training objective follows the factorization in Eq.~\ref{eq:factorization}:
we first supervise the diffusion backbone to recover ChronoFlow trajectories,
and then align the learned interaction representation with action prediction.

To explicitly enforce interaction-centric dynamics learning, the denoising
network $\epsilon_\theta$ is trained using the standard diffusion objective:
\begin{equation}
\mathcal{L}_{\mathrm{CF}}
=
\mathbb{E}_{k}
\left[
\|
\epsilon -
\epsilon_\theta(\bm{P}^{k}_{t:t+H}, \bm{o}_t, \bm{P}_{:t}, k)
\|^2
\right].
\end{equation}
This ChronoFlow supervision encourages the backbone to model
past-current-future consistent object-gripper interaction trajectories.

For action learning, directly using fully denoised trajectories would require
completing the entire diffusion process for every optimization step.
Considering that partially denoised hidden states already encode rich information~\cite{hu2025videopredictionpolicygeneralist, pai2025mimicvideovideoactionmodelsgeneralizable}, 
we condition the action decoder on \textit{partially denoised} ChronoFlow tokens $\bm{Z}^{k-1}_{t}$ during training.
Additionally, to provide an explicit interaction cue, we reconstruct an \emph{approximate} ChronoFlow trajectory $\tilde{\bm{P}}_{t:t+H}$ from an intermediate noisy diffusion state: 
\begin{equation}
\tilde{\bm{P}}_{t:t+H}
=
\frac{1}{\sqrt{\bar{\alpha}_k}}
\Big(
\bm{P}^{k}_{t:t+H}
-
\sqrt{1-\bar{\alpha}_k}\,
\epsilon_\theta(\bm{P}^{k}_{t:t+H},\,\bm{o}_t,\,\bm{P}_{:t},\,k)
\Big).
\end{equation}
The action decoder takes the partially denoised tokens $\bm{Z}^{k-1}_{t}$ and the approximate trajectory $\tilde{\bm{P}}_{t:t+H}$ as input, and is trained with an MSE loss against ground-truth actions:
\begin{equation}
\mathcal{L}_{\mathrm{action}}
=
\mathbb{E}_{k}
\left[
\|
\mathcal{A}_\phi([\bm{Z}^{k-1}_{t};  \tilde{\bm{P}}_{t:t+H}])
-
\bm{a}_{t:t+H}
\|^2
\right].
\end{equation}
The final objective combines action supervision with ChronoFlow supervision:
\begin{equation}
\mathcal{L}
=
\mathcal{L}_{\mathrm{action}}
+
\lambda_{\mathrm{CF}}\mathcal{L}_{\mathrm{CF}}.
\end{equation}
In this formulation, the ChronoFlow loss shapes the diffusion backbone into a
unified spatiotemporal interaction model, while the action loss aligns the
learned representation with task-level control objectives.

\boldparagraph{Inference.}
During inference, historical ChronoFlow trajectories $\bm{P}_{:t}$ are obtained asynchronously using a 3D tracker~\cite{zhang2025tapip3dtrackingpointpersistent}. Starting from Gaussian noise $\bm{P}^{K}_{t:t+H}$, the ChronoFlow encoder and diffusion backbone iteratively produce clean interaction tokens $\bm{Z}^{0}_{t}$. The ChronoFlow and action decoders then generate the future trajectory $\hat{\bm{P}}_{t:t+H}$ and actions $\hat{\bm{a}}_{t:t+H}$.

\section{Experiments}
\label{sec:experiments}

\noindent\textbf{Experiment Overview.}
We evaluate ChronoFlow-Policy in both simulation and real-world manipulation.
Simulation benchmarks provide controlled comparisons against action-only, scene-flow, and object-pose supervision, highlighting the effect of interaction-centric future modeling.
Real-world experiments further test the full past-current-future formulation in long-horizon, deformable, and non-Markovian tasks.

\cref{tab:baseline_comparison} summarizes the compared policy paradigms by supervision signal, history usage, future modeling, and overall gains.
Action-only policies such as DP3~\cite{ze20243ddiffusionpolicygeneralizable} and RISE~\cite{wang2024rise} rely solely on action supervision; history-aware variants: HistRISE~\cite{chen2025historyawarevisuomotorpolicylearning} add past observations without future prediction; 3D-FDP~\cite{noh20253dflowdiffusionpolicy} and MBA~\cite{su2025motionactiondiffusingobject} introduce future supervision through dense scene flow or object pose trajectories.
In contrast, ChronoFlow-Policy (\textbf{CFP}) uses interaction-centric object--gripper ChronoFlow as the future prediction target, and the full model further incorporates historical ChronoFlow tokens.
The \textbf{Gain} column reports absolute improvement over the corresponding base policy: DP3 in simulation and RISE in real-world experiments.

\begin{table}[t]
\centering
\caption{Comparison of methods by supervision, history, and future modeling.}
\label{tab:baseline_comparison}
\begingroup
\fontsize{8pt}{9pt}\selectfont
\setlength{\tabcolsep}{3.0pt}
\renewcommand{\arraystretch}{1.08}
\begin{tabular}{@{}llccc@{}}
\toprule
\textbf{Method}
& \textbf{Supervision}
& \textbf{History}
& \textbf{Future}
& \textbf{Gain} (Sim. / Real) \\
\midrule

DP3 / RISE
& Action-only
& $\times$
& $\times$
& -- / -- \\

HistDP3 / HistRISE
& Action-only
& $\checkmark$
& $\times$
& +2.2 / +25.8 \\

3D-FDP
& Dense scene flow
& $\times$
& $\checkmark$
& +4.0 / +6.9 \\

MBA
& Object pose traj.
& $\times$
& $\checkmark$
& +15.0 / +20.0$^{\dagger}$ \\

\midrule

CFP \textit{w/o past}
& ChronoFlow
& $\times$
& $\checkmark$
& \textbf{+32.5} / +11.8 \\

\textbf{CFP}
& ChronoFlow
& $\checkmark$
& $\checkmark$
& \textbf{+32.5 / +35.3} \\

\bottomrule
\end{tabular}

\begin{minipage}{0.82\textwidth}
\scriptsize
$^{\dagger}$ MBA real-world gain is computed only on \textit{Prepare Breakfast}.
\end{minipage}
\endgroup
\end{table}

\subsection{Simulation Experiments}

\boldparagraph{Benchmarks.}
\begin{itemize}
    \item \textbf{MetaWorld}~\cite{yu2021metaworldbenchmarkevaluationmultitask}.
    MetaWorld is a MuJoCo-based manipulation benchmark with a single gripper interacting with rigid and articulated objects, providing a controlled setting that is well suited for modeling gripper-object interactions and extracting interaction-centric keypoints.
    \item \textbf{RoboTwin}~\cite{chen2025robotwin20scalabledata}. 
    RoboTwin 2.0 is a dual-arm simulation framework for bimanual manipulation, offering diverse objects, tasks, and domain randomization, and enabling the study of coordinated interactions between two arms and objects in complex scenes.
\end{itemize}

\boldparagraph{Baselines.}
For simulation evaluations on MetaWorld and RoboTwin 2.0, we compare ChronoFlow-Policy with representative 3D visuomotor policies: 3D Diffusion Policy (\textbf{DP3})~\cite{ze20243ddiffusionpolicygeneralizable}, 3D Flow Diffusion Policy (\textbf{3D-FDP})~\cite{noh20253dflowdiffusionpolicy}, and Motion Before Action (\textbf{MBA})~\cite{su2025motionactiondiffusingobject}.
Since the simulation benchmarks are largely Markovian, CFP \textit{w/o past} isolates the effect of future ChronoFlow supervision under the same Unet backbone.
This comparison highlights the benefit of ChronoFlow over dense scene flow or object pose supervision.
We implement CFP with both Unet1D~\cite{ronneberger2015unetconvolutionalnetworksbiomedical} and DiT~\cite{peebles2023scalablediffusionmodelstransformers} diffusion backbones, denoted as \textbf{CFP (Unet)} and \textbf{CFP (DiT)} in the following tables.

\boldparagraph{Protocols.}
During training, we adopt a unified temporal setting across all experiments: the action prediction horizon is fixed to 8 steps, the observation horizon is 1 step, and during inference only the first 4 predicted action steps are executed in a receding-horizon manner.
Adhering to the protocol established in~\cite{noh20253dflowdiffusionpolicy}, each MetaWorld experiment is conducted across three trials with seed values 0, 1, and 2. 
The policy is evaluated over 20 episodes every 200 training epochs, and the mean of the top 5 success rates is reported. 
The final performance is computed as the mean and standard deviation across the three seeds.
For RoboTwin 2.0, we evaluate the policy over 100 episodes every 500 training epochs and report the mean of the top 3 success rates. 
The large evaluation sample size yields stable results without requiring multiple random seeds.

\subsubsection{Results.}

\begin{table*}[t]
\centering
\caption{Success rates (\%) on \textbf{MetaWorld}.
Each column corresponds to one task.
Results are reported as mean $\pm$ standard deviation over evaluation rollouts.
Task abbreviations denote \textbf{shelf} (\textit{Shelf Place}), 
\textbf{bin} (\textit{Bin Picking}), 
\textbf{box} (\textit{Box Close}), 
\textbf{peg} (\textit{Peg Insert Side}), 
\textbf{hand} (\textit{Hand Insert}), 
\textbf{soccer} (\textit{Soccer}), and 
\textbf{sweep} (\textit{Sweep Into}).
\textbf{Bold}/blue shading indicates the best result and \underline{underline}/light-blue shading indicates the second-best result.}
\begingroup
\fontsize{8pt}{9pt}\selectfont
\setlength{\tabcolsep}{4pt}
\renewcommand{\arraystretch}{1.2}
\begin{tabular}{lcccccccc}
\toprule

\textbf{Method}
& shelf
& bin
& box
& peg
& hand
& soccer
& sweep
& \textbf{Avg.} \\

\midrule

DP3~\cite{ze20243ddiffusionpolicygeneralizable}
& $17 \pm 10$ & $34 \pm 30$ & $42 \pm 3$ & $69 \pm 7$ & $14 \pm 4$ & $18 \pm 3$ & $15 \pm 5$ & $30$ \\

3D-FDP~\cite{noh20253dflowdiffusionpolicy}
& $16 \pm 3$ & $35 \pm 9$ & $56 \pm 7$ & $70 \pm 8$ & $17 \pm 2$ & $19 \pm 4$ & $25 \pm 7$ & $34$ \\

MBA~\cite{su2025motionactiondiffusingobject}
& \secondcell{$\underline{73} \pm 1$} & $54 \pm 23$ & $56 \pm 2$ & \secondcell{$\underline{75} \pm 5$} & $10 \pm 1$ & $27 \pm 3$ & $34 \pm 25$ & $47$ \\

\textbf{CFP (Unet)}
& $63 \pm 5$ & \bestcell{$\textbf{94} \pm 1$} & \secondcell{$\underline{71} \pm 1$} & \bestcell{$\textbf{89} \pm 3$} & \bestcell{$\textbf{67} \pm 1$} & \bestcell{$\textbf{59} \pm 3$} & \secondcell{$\underline{60} \pm 5$} & \bestcell{$\textbf{72}$} \\

\textbf{CFP (DiT)}
& \bestcell{$\textbf{82} \pm 7$} & \secondcell{$\underline{70} \pm 13$} & \bestcell{$\textbf{75} \pm 5$} & $69 \pm 5$ & \secondcell{$\underline{60} \pm 25$} & \secondcell{$\underline{57} \pm 5$} & \bestcell{$\textbf{75} \pm 5$} & \secondcell{\underline{70}} \\

\bottomrule
\end{tabular}
\endgroup
\label{tab:simulation_metaworld}
\end{table*}

\begin{table*}[t]
\centering
\caption{Success rates (\%) on \textbf{RoboTwin 2.0}.
Each column corresponds to one task.
Task abbreviations denote \textbf{beat} (\textit{Beat Block Hammer}),
\textbf{mic} (\textit{Handover Microphone}),
\textbf{bell} (\textit{Click Bell}),
\textbf{card} (\textit{Move Playingcard Away}),
\textbf{cup} (\textit{Place Empty Cup}),
\textbf{clock} (\textit{Click Alarmclock}),
\textbf{bottles} (\textit{Place Dual Bottles}).
\textbf{Bold}/blue shading indicates the best result and \underline{underline}/light-blue shading indicates the second-best result.
}
\begingroup
\fontsize{8pt}{9pt}\selectfont
\setlength{\tabcolsep}{4pt}
\renewcommand{\arraystretch}{1.2}
\begin{tabular}{lcccccccc}
\toprule

\textbf{Method}
& beat
& mic
& bell
& card
& cup
& clock
& bottles
& \textbf{Avg.} \\

\midrule

DP3~\cite{ze20243ddiffusionpolicygeneralizable}
& 39 & 64 & 66 & 32 & 27 & 54 & 18 & 43 \\

3D-FDP~\cite{noh20253dflowdiffusionpolicy}
& 58 & 65 & 80 & \bestcell{\textbf{41}} & 27 & 44 & 16 & 47 \\

MBA~\cite{su2025motionactiondiffusingobject}
& \secondcell{\underline{61}} & 71 & \secondcell{\underline{84}} & 37 & 39 & \secondcell{\underline{71}} & 27 & 56 \\

\textbf{CFP (Unet)}
& \bestcell{\textbf{77}} & \bestcell{\textbf{88}} & 81 & \secondcell{\underline{39}} & \bestcell{\textbf{59}} & 68 & \bestcell{\textbf{51}} & \bestcell{\textbf{66}} \\

\textbf{CFP (DiT)} 
& \secondcell{\underline{61}} & \secondcell{\underline{82}} & \bestcell{\textbf{93}} & \bestcell{\textbf{41}} & \secondcell{\underline{42}} & \bestcell{\textbf{79}} & \secondcell{\underline{28}} & \secondcell{\underline{63}} \\

\bottomrule
\end{tabular}
\endgroup
\label{tab:simulation_robotwin}
\end{table*}

On MetaWorld, our CFP achieves clear improvements over prior methods across all tasks. 
As shown in \cref{tab:simulation_metaworld}, CFP (Unet) attains an average success rate of 72\%, slightly outperforming CFP (DiT) (70\%), substantially outperforming DP3 (30\%), 3D-FDP (34\%), and MBA (47\%). 
The performance gain is particularly pronounced in interaction-intensive tasks such as \textit{Hand-Insert} and \textit{Sweep Into}.
For example, on \textit{Hand-Insert}, CFP (Unet) achieves 67\% compared to 14\% (DP3) and 17\% (3D-FDP), while on \textit{Sweep Into} it reaches 60\% versus 15\% (DP3).
We attribute these gains to the interaction-centric modeling of ChronoFlow. 
Methods such as 3D-FDP operate on full scene point clouds, which may obscure subtle gripper-object contact patterns within global geometry. 
In contrast, MBA relies solely on object pose supervision and lacks explicit gripper modeling, limiting its ability to reason about fine-grained contact transitions, especially in tasks requiring precise insertion or constrained manipulation like \textit{Sweep Into}.
ChronoFlow’s joint modeling of gripper and object keypoints provides a more expressive and localized representation of interaction dynamics.

On RoboTwin 2.0 \cref{tab:simulation_robotwin}, our CFP also demonstrates consistent improvements in dual-arm manipulation. CFP (Unet) achieves an average success rate of 66\%, surpassing DP3 (43\%), 3D-FDP (47\%), and MBA (56\%), slightly overcomes CFP (DiT) (63\%). 
Despite using only a single observation frame, CFP   performs strongly in tasks requiring accurate spatial localization, such as \textit{Beat Block Hammer} and \textit{Click Alarmclock}, where interaction-based modeling enables more precise alignment and timing. 
Furthermore, in long-horizon dual-arm tasks such as \textit{Handover Microphone}, CFP effectively captures the object transition process between two grippers. 
The explicit modeling of gripper-object interaction flow allows the policy to better represent intermediate transfer states, which are difficult to encode using object-only or scene-level representations.

\subsection{Real-World Experiments}
\label{sec:real_world}
In real-world experiments, we aim to answer the following research questions:
\begin{enumerate}
    \item[$\bullet$] \textbf{Q1:} Does history ChronoFlow improve non-Markovian performance?
    \item[$\bullet$] \textbf{Q2:} Does ChronoFlow policy gain benefits from future modeling?
    \item[$\bullet$] \textbf{Q3:} Can ChronoFlow Policy handle deformable object manipulation?
    \item[$\bullet$] \textbf {Q4:} Which design components of ChronoFlow are critical for real-world robustness (ablation)?
\end{enumerate}

\boldparagraph{Platform.}

\begin{wraptable}[7]{r}{0.46\textwidth}
\vspace{-2.5\baselineskip}
\centering
\caption{Inference frequency.}
\label{tab:inference_speed}
\begingroup
\fontsize{8pt}{9pt}\selectfont
\renewcommand{\arraystretch}{1.08}
\setlength{\tabcolsep}{0.5pt}
\begin{adjustbox}{max width=\linewidth}
\begin{tabular}{@{}lr@{}}
\toprule
\textbf{Method}
& \textbf{Avg. Freq.} \\
\midrule
CFP w/o history track
& 12.18 Hz \\
CFP w/ TAPIP3D (Sync.)
& 0.93 Hz \\
CFP w/ TAPIP3D (Async.)
& 5.82 Hz \\
\bottomrule
\end{tabular}
\end{adjustbox}
\endgroup
\vspace{-0.6\baselineskip}
\end{wraptable}
For real-world experiments, we use a Flexiv Rizon robotic arm equipped with a Robotiq 2F-85 gripper. 
Perception is provided by a single fixed top-down RGB-D camera (Intel RealSense D415), which captures the workspace point cloud for 3D perception. 
All real-world experiments are conducted with a 3D policy only, and no additional cameras or 2D visual inputs are used.
For deployment efficiency, we run TAPIP3D~\cite{zhang2025tapip3dtrackingpointpersistent} asynchronously with the policy.
\cref{tab:inference_speed} shows that synchronous tracking reduces the average inference frequency to 0.93 Hz, whereas the asynchronous pipeline reaches 5.82 Hz.
This design allows CFP to use historical ChronoFlow while avoiding 3D tracking as the runtime bottleneck.

\begin{figure*}[t]
  \centering
  \includegraphics[
    width=\textwidth,
  ]{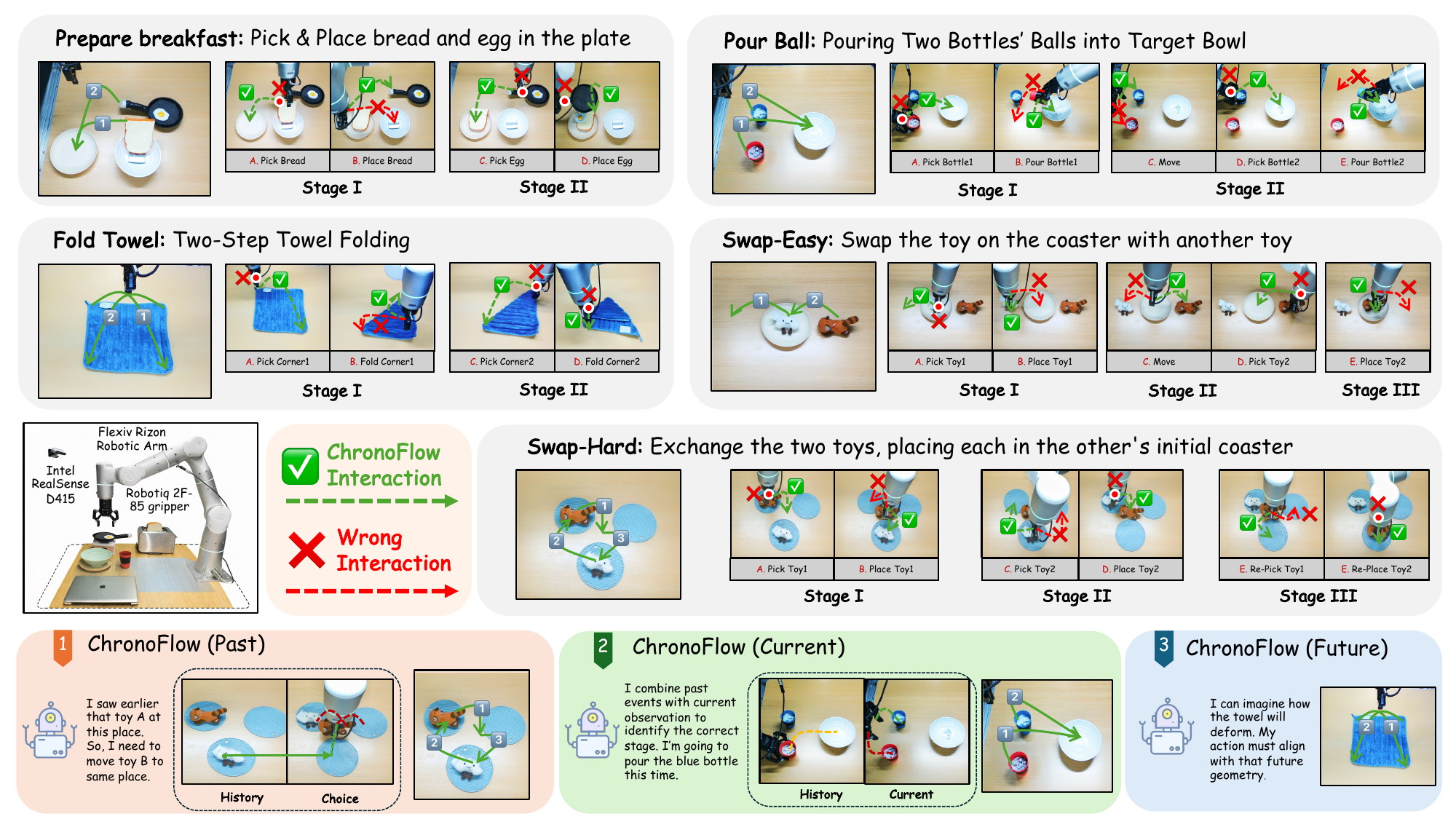}
  \caption{
  \textbf{Real-world manipulation tasks used for evaluation.}
  We consider five real-world tasks,
  including \textit{Prepare Breakfast}, \textit{Pour Ball}, \textit{Fold Towel},
  \textit{Swap-Easy}, and \textit{Swap-Hard}.
  Each task is decomposed into multiple stages.
  }
  \label{fig:real_world_tasks}
\end{figure*}

\boldparagraph{Tasks \& Metrics.}
We select five real-world manipulation tasks for evaluation: \textit{Prepare Breakfast} (long-horizon manipulation), \textit{Fold Towel} (soft-body manipulation), and three non-Markovian tasks including \textit{Swap Easy}, \textit{Swap Hard}, and \textit{Pour Ball}. 
To enable fine-grained analysis of long-horizon behavior, each task is decomposed into a sequence of semantically meaningful stages. Fig.~\ref{fig:real_world_tasks} illustrates the task setups together with their corresponding stage definitions. These staged tasks evaluate the model's ability to remember \textbf{past} interactions, recognize the \textbf{current} manipulation stage, and plan \textbf{future} actions.
Specifically, \textit{Prepare Breakfast} consists of two stages: moving the bread and pouring the egg. 
\textit{Pour Ball} contains two stages corresponding to successfully pouring balls into Bottle1 and Bottle2. 
\textit{Fold Towel} contains two stages representing the first and second folding actions. 
\textit{Swap-Easy} contains three stages: placing Toy1, picking up Toy2, and placing Toy2. 
\textit{Swap-Hard} also contains three stages: placing Toy1 at the intermediate position, swapping Toy2, and finally picking up the intermediate Toy1 to place it at Toy2's original position.

We evaluate all real-world tasks using the one-attempt success rate, defined as whether the task is completed successfully within a single execution without resets. In addition, we record the completion rate of each stage to provide a more detailed analysis of task progress.

\boldparagraph{Baselines.}
We compare ChronoFlow-Policy with representative 3D point-cloud-based baselines:
\textbf{RISE}~\cite{wang2024rise}, History-Aware RISE (\textbf{HistRISE})~\cite{chen2025historyawarevisuomotorpolicylearning},
3D Flow Diffusion Policy (\textbf{3D-FDP})~\cite{noh20253dflowdiffusionpolicy}, and Motion Before Action (\textbf{MBA})~\cite{su2025motionactiondiffusingobject}.
To study the role of historical information \textbf{(Q1)}, we additionally include an ablation variant,
ChronoFlow-Policy without history (\textbf{CFP \textit{w/o past}}), which removes historical keypoints from ChronoFlow while keeping all other components unchanged.
Since MBA relies on object pose trajectories, which are less suitable for history-dependent tasks and deformable-object manipulation, we report MBA on \textit{Prepare Breakfast}, where object-level future motion can be defined more reliably.
For real-world experiments, ChronoFlow-Policy is implemented with a Unet1D diffusion~\cite{pai2025mimicvideovideoactionmodelsgeneralizable} backbone.
For clarity, we denote this variant as \textbf{CFP} in the following tables.
All methods operate on the same 3D point-cloud observations and are trained and evaluated under identical settings whenever applicable.
Gains in \cref{tab:baseline_comparison} are computed relative to RISE, which serves as the action-only base policy with a strong 3D encoder.
CFP \textit{w/o past} isolates the benefit of future ChronoFlow supervision, while full CFP further evaluates whether historical object--gripper flows improve non-Markovian decisions.

\boldparagraph{Protocols.}
For each real-world task, we collect 50 expert demonstrations using the haptic device teleoperation~\cite{rh20t} for training. 
All policies are trained on a workstation equipped with an NVIDIA RTX 3090. During evaluation, the workspace configuration is kept identical across all methods, and test initializations are uniformly sampled to ensure consistent coverage.
Each policy is evaluated over 15 trials per task; for the more challenging \textit{Swap-Hard} task, we increase the evaluation to 18 trials.

\subsubsection{Results.}

\begin{table*}[t]
\centering
\caption{Real-world success rates (\%) across five manipulation tasks:
\textbf{Breakfast} (\textit{Prepare Breakfast}), 
\textbf{Towel} (\textit{Fold Towel}), 
\textbf{Pour} (\textit{Pour Ball}), 
\textbf{Swap-Easy}, and 
\textbf{Swap-Hard}.
Each task is decomposed into sequential stages (I-III); detailed stage definitions are provided in Fig.~\ref{fig:real_world_tasks}.
\textbf{Bold}/blue shading indicates the best performance and \underline{underline}/light-blue shading indicates the second-best performance for each stage.}
\label{tab:real_world}

\begingroup
\fontsize{8pt}{9pt}\selectfont
\setlength{\tabcolsep}{4pt}
\renewcommand{\arraystretch}{1.2}
\begin{tabular}{lccccccccccccc}
\toprule

\textbf{Task} 
& \multicolumn{2}{c}{\textbf{Breakfast}}
& \multicolumn{2}{c}{\textbf{Towel}}
& \multicolumn{2}{c}{\textbf{Pour}}
& \multicolumn{3}{c}{\textbf{Swap-Easy}}
& \multicolumn{3}{c}{\textbf{Swap-Hard}} \\

\cmidrule(lr){2-3}
\cmidrule(lr){4-5}
\cmidrule(lr){6-7}
\cmidrule(lr){8-10}
\cmidrule(lr){11-13}

\textbf{Stage}
& I & II
& I & II 
& I & II
& I & II & III
& I & II & III \\

\midrule

RISE \cite{wang2024rise}
& 73 & 53
& \secondcell{\underline{80}} & 40
& 80 & 47
& \secondcell{\underline{87}} & 27 & 13
& \secondcell{\underline{89}} & 17 & 11 \\

3D FDP \cite{noh20253dflowdiffusionpolicy}
& 73 & \secondcell{\underline{67}}
& \bestcell{\textbf{87}} & \secondcell{\underline{53}}
& 73 & 47
& \bestcell{\textbf{93}} & 40 & 33
& 83 & 33 & 17 \\

HistRISE \cite{chen2025historyawarevisuomotorpolicylearning}
& \secondcell{\underline{87}} & \secondcell{\underline{67}}
& \bestcell{\textbf{87}} & 47
& 80 & \secondcell{\underline{67}}
& \bestcell{\textbf{93}} & \secondcell{\underline{80}} & \secondcell{\underline{80}}
& \bestcell{\textbf{94}} & \secondcell{\underline{89}} & \secondcell{\underline{56}} \\

\midrule

\textbf{CFP} \textit{w/o past}
& \bestcell{\textbf{93}} & \secondcell{\underline{67}}
& \bestcell{\textbf{87}} & \bestcell{\textbf{87}}
& \bestcell{\textbf{93}} & 60
& \bestcell{\textbf{93}} & 33 & 20
& \bestcell{\textbf{94}} & 22 & 11 \\

\textbf{CFP} \textit{(ours)}
& \bestcell{\textbf{93}} & \bestcell{\textbf{80}}
& \bestcell{\textbf{87}} & \bestcell{\textbf{87}}
& \secondcell{\underline{87}} & \bestcell{\textbf{80}}
& \bestcell{\textbf{93}} & \bestcell{\textbf{93}} & \bestcell{\textbf{93}}
& \bestcell{\textbf{94}} & \bestcell{\textbf{94}} & \bestcell{\textbf{61}} \\

\bottomrule
\end{tabular}
\endgroup
\end{table*}

\begin{figure*}[t]
\centering

\begin{subfigure}[t]{0.56\textwidth}
\centering
\includegraphics[
width=\textwidth,
trim=0cm 7.2cm 12.8cm 0cm,
clip
]{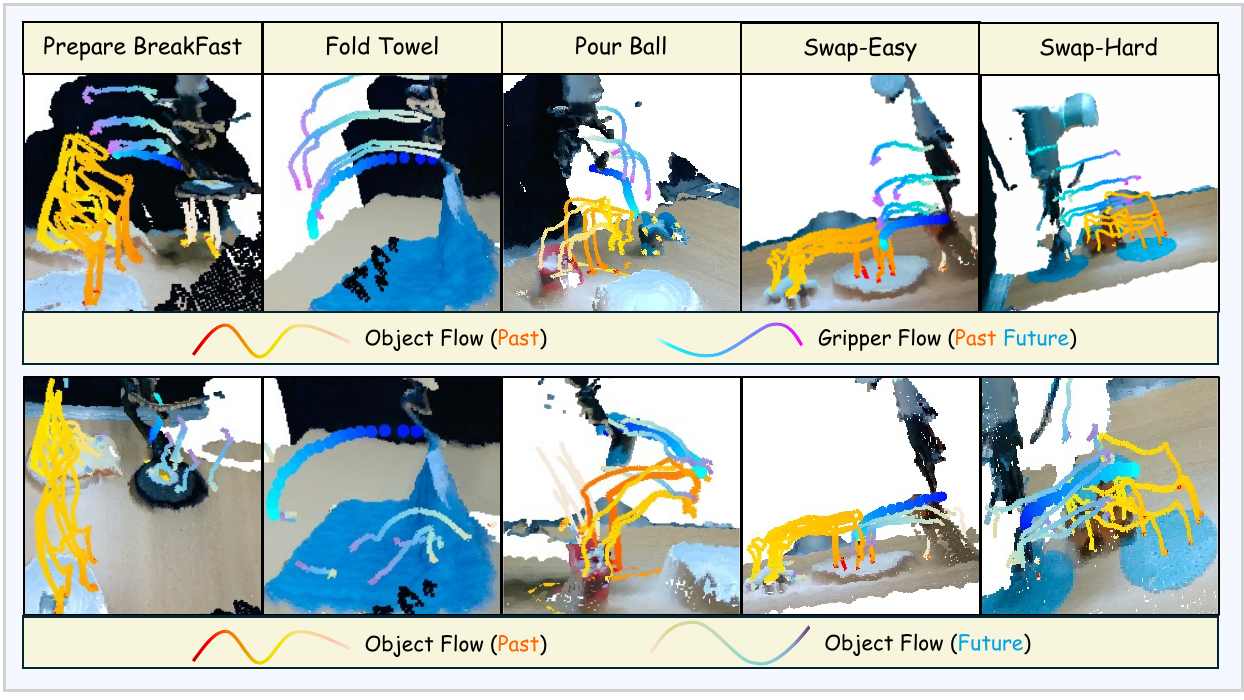}
\caption{}
\label{fig:real}
\end{subfigure}
\hfill
\begin{subfigure}[t]{0.4\textwidth}
\centering
\includegraphics[
width=\textwidth,
]{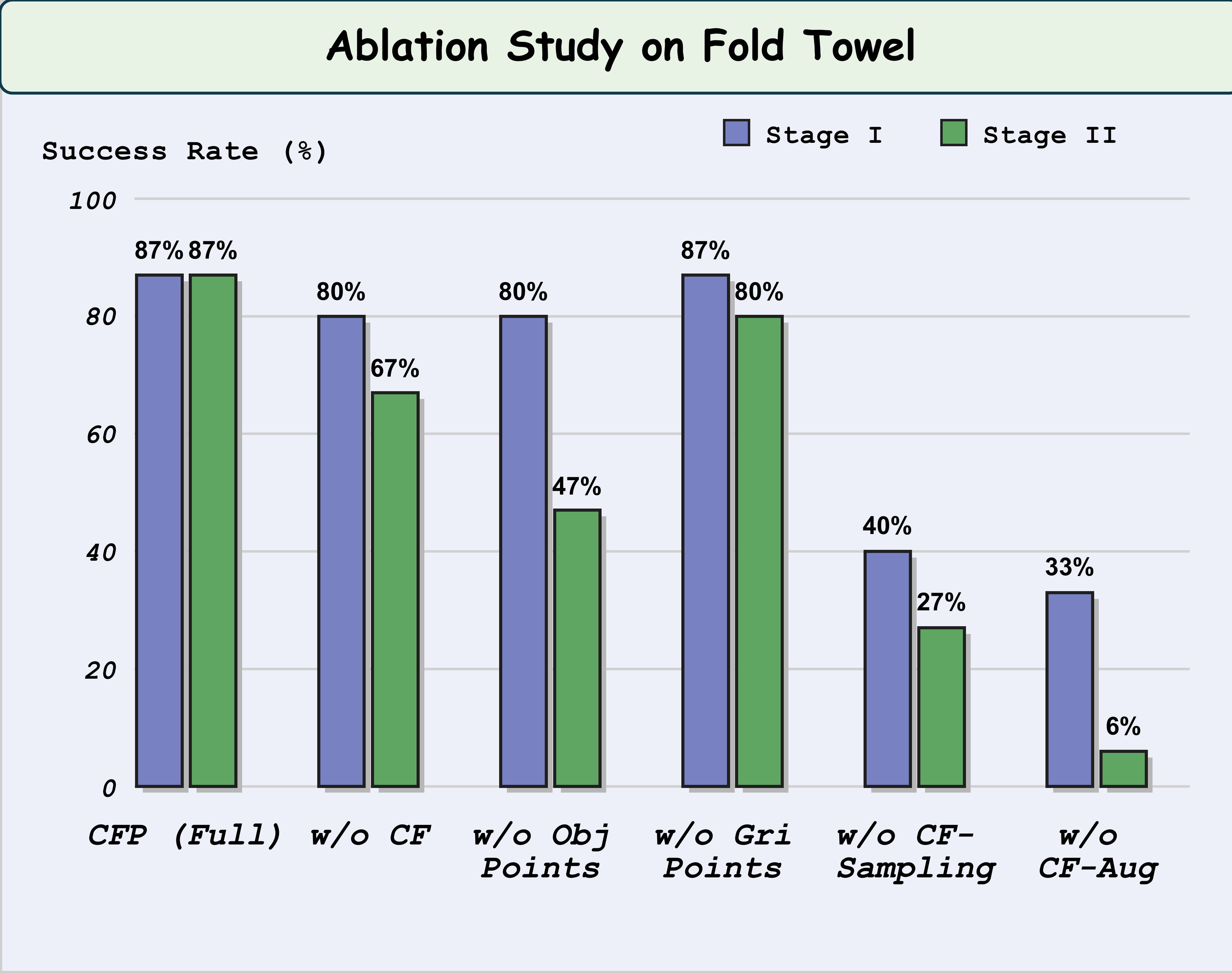}
\caption{}
\label{fig:ablation}
\end{subfigure}
\caption{
\textbf{Left:} Predicted ChronoFlow in real-world tasks: gripper flows (top), object flows (bottom), and historical object flows (yellow/orange).
\textbf{Right:} Ablation on \textit{Fold Towel}.
\textbf{CFP (Full)} denotes the full model.
\textbf{w/o CF} removes ChronoFlow trajectory supervision.
\textbf{w/o ObjPoints} and \textbf{w/o GriPoints} remove object and gripper trajectories from ChronoFlow, respectively.
\textbf{w/o CF-Sampling} disables stochastic subsampling of object keypoints during training.
\textbf{w/o CF-Aug} removes ChronoFlow data augmentation used during training.
}
\end{figure*}

Fig.~\ref{fig:real_world_tasks} summarizes the real-world manipulation results,
while \cref{fig:real} visualizes the interaction-centric keypoint flows predicted by ChronoFlow Policy during execution.
We analyze the performance from three aspects: resolving non-Markovian dependencies (Q1), improving long-horizon task execution (Q2), and handling deformable object manipulation (Q3).

\boldparagraph{Modeling historical interaction improves non-Markovian manipulation (Q1).}
Tasks such as \textit{Swap-Easy} and \textit{Swap-Hard} require decisions that depend on earlier interaction states rather than solely on the current observation.
Methods without explicit history modeling often perform well at the beginning but fail to maintain task progress in later stages.
For example, in \textit{Swap-Easy}, RISE achieves 87\% success in Stage~I but drops sharply to 27\% and 13\% in Stages~II and III.
A similar pattern appears for 3D-FDP (93\% → 40\% → 33\%) and CFP \textit{w/o past} (93\% → 33\% → 20\%).
The same phenomenon is observed in \textit{Swap-Hard}.
In contrast, CFP maintains strong performance across all stages (93\%, 93\%, 93\% in \textit{Swap-Easy} and 94\%, 94\%, 61\% in \textit{Swap-Hard}), showing that historical ChronoFlow effectively resolves non-Markovian dependencies.

\begin{wraptable}[7]{r}{0.36\textwidth}
\vspace{-3.0\baselineskip}
\begin{center}
  \captionsetup{font=footnotesize,skip=2pt,justification=centering,singlelinecheck=false}
  \caption[Comparison with MBA.]{\protect\\[-0.2em]Comparison with MBA.}
  \label{tab:mba_real_world_comp}
  \begingroup
    \fontsize{8pt}{9pt}\selectfont
    \renewcommand{\arraystretch}{1.02}
    \setlength{\tabcolsep}{2.6pt}
    \begin{tabular}{@{}lcc@{}}
      \toprule
      \multirow{2}{*}{\textbf{Method}}
      & \multicolumn{2}{c}{\textbf{Breakfast}} \\
      \cmidrule(lr){2-3}
      & \textbf{Stage I}
      & \textbf{Stage II} \\
      \midrule
      MBA
      & 87\%
      & 73\% \\
      CFP
      & \bestcell{\textbf{93}}\%
      & \bestcell{\textbf{80}}\% \\
      \bottomrule
    \end{tabular}
  \endgroup
\end{center}
\end{wraptable}

\boldparagraph{Future interaction modeling improves long-horizon tasks (Q2).}
On traditional long-horizon tasks such as \textit{Prepare Breakfast}, CFP achieves consistently strong results across both stages.
As shown in \cref{tab:real_world} and \cref{tab:mba_real_world_comp}, our method CFP reaches 93\% success in Stage~I and improves to 80\% in Stage~II, outperforming RISE (73\%, 53\%), 3D-FDP (73\%, 67\%) and MBA (87\%, 73\%).
These results show that joint past-current-future interaction modeling improves temporal reasoning, multi-step planning, and task progression.

\boldparagraph{ChronoFlow Policy effectively handles deformable object manipulation (Q3).}
The \textit{Fold Towel} task highlights the need to model deformable object dynamics.
After the first fold, the towel exhibits highly variable geometry, making Stage~II difficult for methods with limited object representations.
RISE drops from 80\% to 40\%, often failing to localize or grasp the folded towel.
MBA relies on object pose trajectories, but pose estimation for deformable objects is inherently ambiguous and can introduce substantial noise.
In contrast, CFP models interaction-centric keypoint flows, which better preserve towel geometry under deformation.
As a result, CFP achieves 87\% success in both stages and substantially outperforms prior baselines in Stage~II.

\subsubsection{Ablations.} \textbf{\textit{Ablation studies confirm the importance of ChronoFlow design components (Q4).}}
To assess the key design components of CFP, we conduct five ablation experiments on \textit{Fold Towel}, as shown in Fig.~\ref{fig:ablation}.

First, we separately ablate ChronoFlow supervision and flow inputs.
Without ChronoFlow supervision (\emph{w/o ChronoFlow}), Stage~II performance drops from 87\% to 67\%.
Removing object flows (\emph{w/o ObjPoints}) causes a larger Stage~II drop (87\% $\rightarrow$ 47\%) than removing gripper flows (\emph{w/o GriPoints}, 87\% $\rightarrow$ 80\%), indicating that object flows are the primary signal for deformable-object tracking while gripper flows provide complementary cues.

Second, disabling keypoint sampling (\emph{w/o Sampling}) degrades both stages (Stage~I: 87\% $\rightarrow$ 40\%; Stage~II: 87\% $\rightarrow$ 27\%), showing that random keypoint sampling improves spatial coverage and reduces overfitting to fixed tracks.

To improve robustness to viewpoint changes and sensing noise in real-world deployment, our full model applies the same spatial transformation jitter to both ChronoFlow trajectories and point clouds. Third, We disable this augmentation in \emph{w/o CF-Aug}, causing Stage~II success to drop from 87\% to 6\%, suggesting that spatial jittering is critical for stable real-world manipulation.

\section{Discussion and Limitations}
\label{sec:discussion}

ChronoFlow represents temporally consistent object--gripper interactions with sparse 3D trajectories and jointly optimizes flow prediction and action generation, following the world-action modeling principle of learning future dynamics together with control.
Although trajectories extracted by segmentation and 3D tracking contain noise, drift, and missing keypoints, we train with the same imperfect pipeline used at inference, together with random keypoint sampling, dropout, and history truncation.
This encourages the policy to capture stable motion trends rather than rely on individual tracks.
However, severe occlusion or persistent tracking failures may still degrade performance, motivating more robust and uncertainty-aware interaction tracking.

\section*{Acknowledgments}
\vspace{-5pt}
This work was supported by 
the Fundamental and Interdisciplinary Disciplines Breakthrough Plan of the Ministry of Education of China; National Natural Science Foundation of China (No. 62506232); Science and Technology Major Project of Jiangsu Province (No. BG2024041); Shanghai Committee of Science and Technology (Yangfan: No. 24YF2722000; No. 24511103200).

Lixin Yang is the corresponding author and is affiliated with the School of Artificial Intelligence, Shanghai Jiao Tong University, Shanghai, China, 200230.

\bibliographystyle{splncs04}
\bibliography{main}

@String(CVPR  = {IEEE Conf. Comput. Vis. Pattern Recog.})

@String(ECCV  = {Eur. Conf. Comput. Vis.})

@String(NeurIPS = {Adv. Neural Inform. Process. Syst.})

@String(CVPR  = {CVPR})

@String(ECCV  = {ECCV})

@String(NeurIPS = {NeurIPS})

@inproceedings{peebles2023scalablediffusionmodelstransformers,
  title={Scalable diffusion models with transformers},
  author={Peebles, William and Xie, Saining},
  booktitle={Proceedings of the IEEE/CVF international conference on computer vision},
  pages={4195--4205},
  year={2023}
}

@inproceedings{ronneberger2015unetconvolutionalnetworksbiomedical,
  title={U-net: Convolutional networks for biomedical image segmentation},
  author={Ronneberger, Olaf and Fischer, Philipp and Brox, Thomas},
  booktitle={International Conference on Medical Image Computing and Computer-Assisted Intervention},
  pages={234--241},
  year={2015},
  organization={Springer}
}

@misc{torne2026memmultiscaleembodiedmemory,
      title={MEM: Multi-Scale Embodied Memory for Vision Language Action Models}, 
      author={Marcel Torne and Karl Pertsch and Homer Walke and Kyle Vedder and Suraj Nair and Brian Ichter and Allen Z. Ren and Haohuan Wang and Jiaming Tang and Kyle Stachowicz and Karan Dhabalia and Michael Equi and Quan Vuong and Jost Tobias Springenberg and Sergey Levine and Chelsea Finn and Danny Driess},
      year={2026},
      journal={arXiv preprint arXiv:2603.03596}
}

@inproceedings{ravi2024sam2segmentimages,
  title={SAM 2: Segment Anything in Images and Videos},
  author={Ravi, Nikhila and Gabeur, Valentin and Hu, Yuan-Ting and Hu, Ronghang and Ryali, Chaitanya and Ma, Tengyu and Khedr, Haitham and R{\"a}dle, Roman and Rolland, Chloe and Gustafson, Laura and others},
  booktitle={International Conference on Learning Representations},
  year={2025}
}

@inproceedings{ze20243ddiffusionpolicygeneralizable,
  author       = {Yanjie Ze and
                  Gu Zhang and
                  Kangning Zhang and
                  Chenyuan Hu and
                  Muhan Wang and
                  Huazhe Xu},
  title        = {3D Diffusion Policy: Generalizable Visuomotor Policy Learning via
                  Simple 3D Representations},
  booktitle    = {Robotics: Science and Systems},
  year         = {2024}
}

@inproceedings{wang2024rise,
    title     = {RISE: 3D Perception Makes Real-World Robot Imitation Simple and Effective},
    author    = {Wang, Chenxi and Fang, Hongjie and Fang, Hao-Shu and Lu, Cewu},
    booktitle = {IEEE/RSJ International Conference on Intelligent Robots and Systems}, 
    year      = {2024},
    pages     = {2870-2877}
}

@inproceedings{xiao2025spatialtrackerv23dpointtracking,
  title={Spatialtrackerv2: Advancing 3d point tracking with explicit camera motion},
  author={Xiao, Yuxi and Wang, Jianyuan and Xue, Nan and Karaev, Nikita and Makarov, Yuri and Kang, Bingyi and Zhu, Xing and Bao, Hujun and Shen, Yujun and Zhou, Xiaowei},
  booktitle={Proceedings of the IEEE/CVF International Conference on Computer Vision},
  pages={6726--6737},
  year={2025}
}

@article{zhang2025tapip3dtrackingpointpersistent,
  title={TAPIP3D: Tracking Any Point in Persistent 3D Geometry},
  author={Zhang, Bowei and Ke, Lei and Harley, Adam W and Fragkiadaki, Katerina},
  journal={arXiv preprint arXiv:2504.14717},
  year={2025}
}

@article{chen2025historyawarevisuomotorpolicylearning,
      title={History-Aware Visuomotor Policy Learning via Point Tracking}, 
      author={Jingjing Chen and Hongjie Fang and Chenxi Wang and Shiquan Wang and Cewu Lu},
      year={2025},
      journal={arXiv preprint arXiv:2509.17141},
}

@inproceedings{zheng2025tracevlavisualtraceprompting,
  title={TraceVLA: Visual Trace Prompting Enhances Spatial-Temporal Awareness for Generalist Robotic Policies},
  author={Zheng, Ruijie and Liang, Yongyuan and Huang, Shuaiyi and Gao, Jianfeng and Daum{\'e} III, Hal and Kolobov, Andrey and Huang, Furong and Yang, Jianwei},
  booktitle={International Conference on Learning Representations},
  year={2025}
}

@article{su2025motionactiondiffusingobject,
  author       = {Yue Su and
                  Xinyu Zhan and
                  Hongjie Fang and
                  Yong{-}Lu Li and
                  Cewu Lu and
                  Lixin Yang},
  title        = {Motion Before Action: Diffusing Object Motion as Manipulation Condition},
  journal      = {IEEE Robotics and Automation Letters},
  volume       = {10},
  number       = {7},
  pages        = {7428--7435},
  year         = {2025}
}

@inproceedings{hsu2025spotse3posetrajectory,
  title={SPOT: SE(3) pose trajectory diffusion for object-centric manipulation},
  author={Hsu, Cheng-Chun and Wen, Bowen and Xu, Jie and Narang, Yashraj and Wang, Xiaolong and Zhu, Yuke and Biswas, Joydeep and Birchfield, Stan},
  booktitle={IEEE International Conference on Robotics and Automation},
  pages={4853--4860},
  year={2025},
}

@article{noh20253dflowdiffusionpolicy,
  title={3d flow diffusion policy: Visuomotor policy learning via generating flow in 3d space},
  author={Noh, Sangjun and Nam, Dongwoo and Kim, Kangmin and Lee, Geonhyup and Yu, Yeonguk and Kang, Raeyoung and Lee, Kyoobin},
  journal={arXiv preprint arXiv:2509.18676},
  year={2025}
}

@article{huang2026pointworldscaling3dworld,
  title={PointWorld: Scaling 3D World Models for In-The-Wild Robotic Manipulation},
  author={Huang, Wenlong and Chao, Yu-Wei and Mousavian, Arsalan and Liu, Ming-Yu and Fox, Dieter and Mo, Kaichun and Fei-Fei, Li},
  journal={arXiv preprint arXiv:2601.03782},
  year={2026}
}

@inproceedings{hu2025videopredictionpolicygeneralist,
  title={Video Prediction Policy: A Generalist Robot Policy with Predictive Visual Representations},
  author={Hu, Yucheng and Guo, Yanjiang and Wang, Pengchao and Chen, Xiaoyu and Wang, Yen-Jen and Zhang, Jianke and Sreenath, Koushil and Lu, Chaochao and Chen, Jianyu},
  booktitle={International Conference on Machine Learning},
  pages={24328--24346},
  year={2025}
}

@inproceedings{yu2021metaworldbenchmarkevaluationmultitask,
  author       = {Tianhe Yu and
                  Deirdre Quillen and
                  Zhanpeng He and
                  Ryan Julian and
                  Karol Hausman and
                  Chelsea Finn and
                  Sergey Levine},
  title        = {Meta-World: {A} Benchmark and Evaluation for Multi-Task and Meta Reinforcement Learning},
  booktitle    = {Conference on Robot Learning},
  series       = {Proceedings of Machine Learning Research},
  volume       = {100},
  pages        = {1094--1100},
  publisher    = {{PMLR}},
  year         = {2019}
}

@inproceedings{rh20t,
  author       = {Hao-Shu Fang and
                  Hongjie Fang and
                  Zhenyu Tang and
                  Jirong Liu and
                  Chenxi Wang and
                  Junbo Wang and
                  Haoyi Zhu and
                  Cewu Lu},
  title        = {{RH20T:} {A} Comprehensive Robotic Dataset for Learning Diverse Skills
                  in One-Shot},
  booktitle    = {{IEEE} International Conference on Robotics and Automation},
  pages        = {653--660},
  publisher    = {{IEEE}},
  year         = {2024},
}

@article{chen2025robotwin20scalabledata,
  title={Robotwin 2.0: A scalable data generator and benchmark with strong domain randomization for robust bimanual robotic manipulation},
  author={Chen, Tianxing and Chen, Zanxin and Chen, Baijun and Cai, Zijian and Liu, Yibin and Li, Zixuan and Liang, Qiwei and Lin, Xianliang and Ge, Yiheng and Gu, Zhenyu and others},
  journal={arXiv preprint arXiv:2506.18088},
  year={2025}
}

@article{pai2025mimicvideovideoactionmodelsgeneralizable,
  title={mimic-video: Video-action models for generalizable robot control beyond vlas},
  author={Pai, Jonas and Achenbach, Liam and Montesinos, Victoriano and Forrai, Benedek and Mees, Oier and Nava, Elvis},
  journal={arXiv preprint arXiv:2512.15692},
  year={2025}
}

@article{dreamzero2025,
  title={World Action Models are Zero-shot Policies},
  author={Ye, Seonghyeon and Ge, Yunhao and Zheng, Kaiyuan and Gao, Shenyuan and Yu, Sihyun and Kurian, George and Indupuru, Suneel and Tan, You Liang and Zhu, Chuning and Xiang, Jiannan and others},
  journal={arXiv preprint arXiv:2602.15922},
  year={2026}
}

@article{lingbot-va2026,
  title={Causal World Modeling for Robot Control},
  author={Li, Lin and Zhang, Qihang and Luo, Yiming and Yang, Shuai and Wang, Ruilin and Han, Fei and Yu, Mingrui and Gao, Zelin and Xue, Nan and Zhu, Xing and Shen, Yujun and Xu, Yinghao},
  journal={arXiv preprint arXiv:2601.21998},
  year={2026}
}

@inproceedings{huang2022flowformertransformerarchitectureoptical,
  title={Flowformer: A transformer architecture for optical flow},
  author={Huang, Zhaoyang and Shi, Xiaoyu and Zhang, Chao and Wang, Qiang and Cheung, Ka Chun and Qin, Hongwei and Dai, Jifeng and Li, Hongsheng},
  booktitle={European conference on computer vision},
  pages={668--685},
  year={2022},
  organization={Springer}
}

@inproceedings{shi2023flowformermaskedcostvolume,
  title={Flowformer++: Masked cost volume autoencoding for pretraining optical flow estimation},
  author={Shi, Xiaoyu and Huang, Zhaoyang and Li, Dasong and Zhang, Manyuan and Cheung, Ka Chun and See, Simon and Qin, Hongwei and Dai, Jifeng and Li, Hongsheng},
  booktitle={Proceedings of the IEEE/CVF conference on computer vision and pattern recognition},
  pages={1599--1610},
  year={2023}
}

@article{gkanatsios20253dflowmatchactorunified,
  title={3D FlowMatch Actor: Unified 3D Policy for Single-and Dual-Arm Manipulation},
  author={Gkanatsios, Nikolaos and Xu, Jiahe and Bronars, Matthew and Mousavian, Arsalan and Ke, Tsung-Wei and Fragkiadaki, Katerina},
  journal={arXiv preprint arXiv:2508.11002},
  year={2025}
}

@inproceedings{xu2024flowcrossdomainmanipulationinterface,
  author       = {Mengda Xu and
                  Zhenjia Xu and
                  Yinghao Xu and
                  Cheng Chi and
                  Gordon Wetzstein and
                  Manuela Veloso and
                  Shuran Song},
  title        = {Flow as the Cross-Domain Manipulation Interface},
  booktitle    = {Conference on Robot Learning},
  volume       = {270},
  pages        = {2475--2499},
  publisher    = {{PMLR}},
  year         = {2024}
}

@inproceedings{wen2024anypointtrajectorymodelingpolicy,
  author       = {Chuan Wen and
                  Xingyu Lin and
                  John Ian Reyes So and
                  Kai Chen and
                  Qi Dou and
                  Yang Gao and
                  Pieter Abbeel},
  title        = {Any-point Trajectory Modeling for Policy Learning},
  booktitle    = {Robotics: Science and Systems},
  year         = {2024}
}

@inproceedings{bharadhwaj2024track2actpredictingpointtracks,
  title={Track2act: Predicting point tracks from internet videos enables generalizable robot manipulation},
  author={Bharadhwaj, Homanga and Mottaghi, Roozbeh and Gupta, Abhinav and Tulsiani, Shubham},
  booktitle={European Conference on Computer Vision},
  pages={306--324},
  year={2024},
  organization={Springer}
}

@article{su2025freqpolicyefficientflowbasedvisuomotor,
  title={Freqpolicy: Efficient flow-based visuomotor policy via frequency consistency},
  author={Su, Yifei and Liu, Ning and Chen, Dong and Zhao, Zhen and Wu, Kun and Li, Meng and Xu, Zhiyuan and Che, Zhengping and Tang, Jian},
  journal={arXiv preprint arXiv:2506.08822},
  year={2025}
}

@inproceedings{eisner2024flowbot3dlearning3darticulation,
  author       = {Ben Eisner and
                  Harry Zhang and
                  David Held},
  title        = {FlowBot3D: Learning 3D Articulation Flow to Manipulate Articulated Objects},
  booktitle    = {Robotics: Science and Systems},
  year         = {2022}
}

@inproceedings{fang2025sam2actintegratingvisualfoundation,
  title={SAM2Act: Integrating Visual Foundation Model with A Memory Architecture for Robotic Manipulation},
  author={Fang, Haoquan and Grotz, Markus and Pumacay, Wilbert and Wang, Yi Ru and Fox, Dieter and Krishna, Ranjay and Duan, Jiafei},
  booktitle={International Conference on Machine Learning},
  year={2025}
}

@inproceedings{tang2025adaptivekeyframesamplinglong,
  title={Adaptive keyframe sampling for long video understanding},
  author={Tang, Xi and Qiu, Jihao and Xie, Lingxi and Tian, Yunjie and Jiao, Jianbin and Ye, Qixiang},
  booktitle={Proceedings of the Computer Vision and Pattern Recognition Conference},
  pages={29118--29128},
  year={2025}
}

@inproceedings{yu2025framevoyagerlearningqueryframes,
  title={Frame-voyager: Learning to query frames for video large language models},
  author={Yu, Sicheng and Jin, Chengkai and Wang, Huanyu and Chen, Zhenghao and Jin, Sheng and Zuo, Zhongrong and Xu, Xiaolei and Sun, Zhenbang and Zhang, Bingni and Wu, Jiawei and others},
  booktitle={International Conference on Learning Representations},
  year={2025}
}

@inproceedings{jin2024videolavitunifiedvideolanguagepretraining,
  title={Video-LaVIT: unified video-language pre-training with decoupled visual-motional tokenization},
  author={Jin, Yang and Sun, Zhicheng and Xu, Kun and Chen, Liwei and Jiang, Hao and Huang, Quzhe and Song, Chengru and Liu, Yuliang and Zhang, Di and Song, Yang and others},
  booktitle={International Conference on Machine Learning},
  pages={22185--22209},
  year={2024}
}

@article{shi2026memoryvlaperceptualcognitivememoryvisionlanguageaction,
  title={Memoryvla: Perceptual-cognitive memory in vision-language-action models for robotic manipulation},
  author={Shi, Hao and Xie, Bin and Liu, Yingfei and Sun, Lin and Liu, Fengrong and Wang, Tiancai and Zhou, Erjin and Fan, Haoqiang and Zhang, Xiangyu and Huang, Gao},
  journal={arXiv preprint arXiv:2508.19236},
  year={2025}
}

@inproceedings{SlowFast-LLaVA-1.5,
  title={SlowFast-LLaVA-1.5: A Family of Token-Efficient Video Large Language Models for Long-Form Video Understanding},
  author={Xu, Mingze and Gao, Mingfei and Li, Shiyu and Lu, Jiasen and Gan, Zhe and Lai, Zhengfeng and Cao, Meng and Kang, Kai and Yang, Yinfei and Dehghan, Afshin},
  booktitle={Conference on Language Modeling},
  year={2025}
}

@article{li2025cronusvlaefficientrobustmanipulation,
  title={CronusVLA: Towards Efficient and Robust Manipulation via Multi-Frame Vision-Language-Action Modeling},
  author={Li, Hao and Yang, Shuai and Chen, Yilun and Chen, Xinyi and Yang, Xiaoda and Tian, Yang and Wang, Hanqing and Wang, Tai and Lin, Dahua and Zhao, Feng and others},
  journal={arXiv preprint arXiv:2506.19816},
  year={2025}
}

@inproceedings{diffusion_forcing,
  author       = {Boyuan Chen and
                  Diego Marti Monso and
                  Yilun Du and
                  Max Simchowitz and
                  Russ Tedrake and
                  Vincent Sitzmann},
  title        = {Diffusion Forcing: Next-token Prediction Meets Full-Sequence Diffusion},
  booktitle    = {NeurIPS},
  pages = {24081--24125},
  year         = {2024}
}

@inproceedings{ptp,
  title={Learning Long-Context Diffusion Policies via Past-Token Prediction},
  author={Torne, Marcel and Tang, Andy and Liu, Yuejiang and Finn, Chelsea},
  booktitle={CoRL},
  year={2025}
}

@article{memoryvla,
  title={MemoryVLA: Perceptual-Cognitive Memory in Vision-Language-Action Models for Robotic Manipulation},
  author={Shi, Hao and Xie, Bin and Liu, Yingfei and Sun, Lin and Liu, Fengrong and Wang, Tiancai and Zhou, Erjin and Fan, Haoqiang and Zhang, Xiangyu and Huang, Gao},
  journal={arXiv preprint arXiv:2508.19236},
  year={2025}
}

@inproceedings{zhu2025uwm,
    author    = {Zhu, Chuning and Yu, Raymond and Feng, Siyuan and Burchfiel, Benjamin and Shah, Paarth and Gupta, Abhishek},
    title     = {Unified World Models: Coupling Video and Action Diffusion for Pretraining on Large Robotic Datasets},
    booktitle = {Proceedings of Robotics: Science and Systems (RSS)},
    year      = {2025},
}

@inproceedings{zhang2026d4rt,
  title={Efficiently Reconstructing Dynamic Scenes One D4RT at a Time},
  author={Zhang, Chuhan and Le Moing, Guillaume and Koppula, Skanda and Rocco, Ignacio and Momeni, Liliane and Xie, Junyu and Sun, Shuyang and Sukthankar, Rahul and Barral, Jo{\"e}lle K. and Hadsell, Raia and Ghahramani, Zoubin and Zisserman, Andrew and Zhang, Junlin and Sajjadi, Mehdi S. M.},
  booktitle={CVPR},
  year={2026}
}

@inproceedings{wang2026lamp,
  title={{LaMP}: Learning Vision-Language-Action Policy with 3D Scene Flow as Latent Motion Prior},
  author={Wang, Xinkai and Wang, Chenyi and Xu, Yifu and Ye, Mingzhe and Zhang, Fucheng and Tian, Jialin and Zhan, Xinyu and Zhu, Lifeng and Lu, Cewu and Yang, Lixin},
  booktitle={European Conference on Computer Vision (ECCV)},
  year={2026}
}
\appendix
\section*{Appendix}
\section{ChronoFlow Implementation Details}
This section details the construction of \textbf{ChronoFlow}, including how gripper and object keypoints are defined, how their temporal flows are obtained, and how historical and future ChronoFlow trajectories are organized.

\boldparagraph{ChronoFlow keypoints.}
ChronoFlow represents the gripper-object interaction state at each timestep $t$ using sparse 3D keypoints defined on both the robot gripper and task-relevant objects.
We denote the gripper keypoints as $\bm{P}^g_t \in \mathbb{R}^{N_g \times 3}$ and the object keypoints as $\bm{P}^o_t \in \mathbb{R}^{N_o \times 3}$, where $N_g$ and $N_o$ are the numbers of gripper and object keypoints, respectively.
Their union $\bm{P}_t = \bm{P}^g_t \cup \bm{P}^o_t$ forms a compact abstraction of the physical interaction state.

\boldparagraph{Gripper keypoints flow.}
We predefine $N_g$ canonical keypoints on the gripper geometry, rigidly attached to the tool center point (TCP).
Given the TCP pose at each timestep, we compute the 3D positions of all gripper keypoints via rigid transformation, producing temporally consistent gripper trajectories in the world coordinate frame.

\boldparagraph{Object keypoints flow.}
For manipulated objects, we segment task-relevant objects from the initial observation, e.g., using 3D SAM-2~\cite{ravi2024sam2segmentimages}, and apply Farthest Point Sampling (FPS) to obtain candidate keypoints on each object.
We then track these points over time using a 3D point tracker, e.g., TAPIP3D~\cite{zhang2025tapip3dtrackingpointpersistent}, to obtain temporally consistent object-centric trajectories.
In practice, instead of using all tracked candidates, we randomly sample $N_o$ object keypoints per sequence for training, which reduces overfitting to specific spatial locations and improves generalization across interaction configurations.

\boldparagraph{Historical ChronoFlow trajectory.}
Given a history horizon $h_p$, we construct the historical ChronoFlow trajectory as
$\bm{P}_{t-h_p:t-1} = \{\bm{P}_{t-h_p}, \ldots, \bm{P}_{t-1}\}$, where
$\bm{P}^g_{t-h_p:t-1} \in \mathbb{R}^{h_p \times N_g \times 3}$ and
$\bm{P}^o_{t-h_p:t-1} \in \mathbb{R}^{h_p \times N_o \times 3}$.
It contains both gripper and object keypoint flows over the past window.
For conciseness, we use $\bm{P}_{:t}=\{\bm{P}_\tau\}_{\tau<t}$ to denote the historical ChronoFlow trajectories constructed from past observations.

\boldparagraph{Future ChronoFlow trajectory.}
Following the notation in the main paper, the policy predicts a horizon-length ChronoFlow trajectory
$\bm{P}_{t:t+H}$, which describes the future evolution of gripper-object interactions starting from timestep $t$.
Its gripper and object components are denoted as
$\bm{P}^g_{t:t+H} \in \mathbb{R}^{H \times N_g \times 3}$ and
$\bm{P}^o_{t:t+H} \in \mathbb{R}^{H \times N_o \times 3}$, respectively.
Equivalently, the full trajectory is represented as
$\bm{P}_{t:t+H}\in\mathbb{R}^{(N_g+N_o)\times H\times 3}$.

\section{ChronoFlow Policy Implementation Details}
\label{sec:cfp-impl}
This section provides implementation details of \textbf{ChronoFlow Policy}, including the current observation encoder, and the conditioning mechanisms used in different diffusion backbones, i.e., Unet1D~\cite{ronneberger2015unetconvolutionalnetworksbiomedical} and DiT~\cite{peebles2023scalablediffusionmodelstransformers}.

\subsection{Current Observation Encoder}
\label{sec:cfp-obs-encoder}

\boldparagraph{Observation.}
ChronoFlow Policy conditions on the current observation
$\bm{o}_t=(\bm{p}_t,\bm{q}_t)$ at decision time $t$.
Here $\bm{p}_t\in\mathbb{R}^{N\times 6}$ denotes the scene-level point cloud, where each point contains 3D coordinates and RGB values.
When available, $\bm{q}_t\in\mathbb{R}^{d_q}$ denotes the proprioceptive state, such as joint positions or end-effector pose.
The observation encoder maps $\bm{o}_t$ to a global feature vector
$\bm{c}_t = E_{\mathrm{obs}}(\bm{o}_t)\in\mathbb{R}^{D_{\mathrm{obs}}}$, which conditions the diffusion backbone.

\boldparagraph{Point cloud branch.}
We use different point-cloud encoders in simulation and real-world settings.
In simulation, we adopt a PointNet-style point-cloud encoder composed of shared per-point MLP layers with widths \texttt{[64, 128, 256, 512]}, followed by symmetric max pooling over points and a final linear projection to produce the compact point-cloud embedding $\bm{z}^{\mathrm{pc}}_t \in \mathbb{R}^{D_{\mathrm{pc}}}$.
In real-robot experiments, the raw point cloud is noisier and contains substantially more points than in simulation.
Therefore, we use a sparse 3D encoder to efficiently extract robust point features from dense point clouds.

\boldparagraph{State branch.}
When proprioception is available, we encode $\bm{q}_t\in\mathbb{R}^{d_q}$ using a lightweight MLP with ReLU activations.
In all simulation experiments, we set \texttt{state\_mlp\_size=(64, 64)}, yielding a two-layer MLP
$d_q\!\rightarrow\!64\!\rightarrow\!64$ and producing a state embedding $\bm{z}^{q}_t\in\mathbb{R}^{64}$.
In real-robot experiments, proprioceptive state is omitted and the observation feature is extracted from the point cloud only.

\boldparagraph{Observation feature fusion.}
When both point-cloud and proprioceptive features are available, we concatenate $\bm{z}^{\mathrm{pc}}_t$ and $\bm{z}^{q}_t$ along the channel dimension and apply a final linear layer.
The resulting feature $\bm{c}_t\in\mathbb{R}^{D_{\mathrm{obs}}}$ is used as the observation condition for ChronoFlow diffusion.
We set $D_{\mathrm{obs}}=128$ in our implementation.

\subsection{Conditioning and Diffusion Backbone}
\label{sec:cfp-conditioning}

We implement ChronoFlow Policy with diffusion backbones, including Unet1D~\cite{ronneberger2015unetconvolutionalnetworksbiomedical} and DiT~\cite{peebles2023scalablediffusionmodelstransformers}.
Since our simulation benchmarks are approximately Markovian while real-world manipulation exhibits stronger non-Markovian dependencies, we use different conditioning signals in these two settings.

\boldparagraph{ChronoFlow diffusion target.}
At each timestep $t$, ChronoFlow represents the interaction state as
$\bm{P}_t=\bm{P}^g_t\cup\bm{P}^o_t\in\mathbb{R}^{N\times 3}$, where $N=N_g+N_o$.
For a prediction horizon $H$, the diffusion model predicts the ChronoFlow trajectory
$\bm{P}_{t:t+H}\in\mathbb{R}^{N\times H\times 3}$.
We apply the standard forward diffusion process to $\bm{P}_{t:t+H}$ and denote the noisy trajectory at diffusion step $k$ as $\bm{P}^{k}_{t:t+H}$.
The denoising network predicts the diffusion target, e.g., noise $\epsilon$ under the DDPM objective, conditioned on the current observation and historical ChronoFlow.

\boldparagraph{Simulation setting.}
In simulation, tasks are approximately Markovian and the current interaction state is sufficient for decision making.
Therefore, we inject the current ChronoFlow state $\bm{P}_t$ as an explicit trajectory condition.
Specifically, we repeat $\bm{P}_t$ along the prediction horizon to obtain
$\widetilde{\bm{P}}_{t:t+H}=\textsc{Repeat}(\bm{P}_t)$ and concatenate it with the noisy ChronoFlow trajectory:
\begin{equation}
\bm{S}^{k}_{t:t+H}
=
\left[
\widetilde{\bm{P}}_{t:t+H};
\bm{P}^{k}_{t:t+H}
\right]
\in\mathbb{R}^{H\times N\times 6}.
\end{equation}
The diffusion backbone takes $\bm{S}^{k}_{t:t+H}$, the diffusion timestep $k$, and the observation feature
$\bm{c}_t=E_{\mathrm{obs}}(\bm{o}_t)$ as input, and predicts the denoising target for the noisy ChronoFlow trajectory.

\boldparagraph{Real-world setting.}
In real-world long-horizon manipulation, we run the 3D point tracker asynchronously with the policy for efficiency.
As a result, the policy may not have access to the up-to-date current ChronoFlow state $\bm{P}_t$ at inference time.
Therefore, in real-robot experiments, we do not use $\bm{P}_t$ as an explicit repeated trajectory condition.
Instead, the model denoises the future ChronoFlow trajectory from Gaussian noise while conditioning on both the historical ChronoFlow $\bm{P}_{:t}$ and the current point-cloud observation feature.

Concretely, at each diffusion step $k$, the ChronoFlow encoder jointly encodes $\bm{P}_{:t}$ and $\bm{P}^{k}_{t:t+H}$ into tokens $\bm{Z}^{k}_{t}$.
The observation encoder extracts the current point-cloud feature $\bm{c}_t=E_{\mathrm{obs}}(\bm{p}_t)$.
The diffusion backbone then uses $\bm{Z}^{k}_{t}$, $\bm{c}_t$, and the diffusion timestep $k$ as conditioning signals to perform iterative denoising.

\boldparagraph{Action decoding.}
After iterative denoising, the model obtains clean ChronoFlow tokens $\bm{Z}^{0}_{t}$ and decodes them into the predicted future interaction trajectory
$\hat{\bm{P}}_{t:t+H}=D_{\mathrm{CF}}(\bm{Z}^{0}_{t})$.
The action decoder predicts future robot actions from the clean tokens and the decoded ChronoFlow trajectory:
\begin{equation}
\hat{\bm{a}}_{t:t+H}
=
\mathcal{A}_{\phi}
\left(
\left[
\bm{Z}^{0}_{t};
\hat{\bm{P}}_{t:t+H}
\right]
\right).
\end{equation}
During training, action prediction is supervised with an MSE objective, while the diffusion backbone is trained with ChronoFlow denoising supervision.

\section{Baseline Implementation Details}
\label{sec:baseline-impl}

This section summarizes implementation details of the baselines and highlights their differences from \textbf{ChronoFlow Policy}.
To ensure a fair comparison, all baselines use the same point-cloud observation encoder, normalization, diffusion scheduler, and shared module hyperparameters whenever applicable.
We only vary the core modeling choices, including history usage, interaction representation, prediction targets, and training objectives.

\subsection{RISE}
\label{sec:impl-rise}

RISE~\cite{wang2024rise} is an \emph{action-only} diffusion policy.
\textbf{Key differences to ChronoFlow Policy:} it does not predict interaction trajectories, does not perform keypoint/flow diffusion, and does not incorporate an explicit ChronoFlow encoder.
Given a single point-cloud observation at time $t$, RISE first encodes the point cloud to obtain point features, aggregates them with a Transformer readout, and conditions an action diffusion head on the resulting global feature.

\boldparagraph{Shared observation encoder.}
RISE uses the same point-cloud observation encoder as ChronoFlow Policy to extract point features from the current point cloud.

\boldparagraph{Transformer aggregation.}
Point features are processed by a Transformer with $\texttt{hidden\_dim}=512$, $\texttt{nheads}=8$, $\texttt{n\_encoder\_layers}=4$, $\texttt{n\_decoder\_layers}=1$, $\texttt{dim\_feedforward}=2048$, and dropout $0.1$.
A learnable readout token is appended, and the readout feature $\bm{r}_t\in\mathbb{R}^{512}$ is used as the global condition.

\boldparagraph{Action diffusion head.}
The action head is a conditional Unet1D diffusion model, implemented as \texttt{DiffusionUNetPolicy}.
It predicts $\texttt{num\_action}=20$ action steps with per-step dimension $\texttt{action\_dim}=10$.
RISE is trained and evaluated with \textbf{action diffusion only}.

\subsection{HistRISE}
\label{sec:impl-histrise}

HistRISE~\cite{chen2025historyawarevisuomotorpolicylearning} is an \emph{action-only} diffusion policy that augments RISE with point-track history features.
\textbf{Key differences to ChronoFlow Policy:} HistRISE does not predict future interaction trajectories, has no ChronoFlow supervision, and does not perform diffusion over interaction-centric keypoint representations.
It only trains an action diffusion head.

\boldparagraph{Shared observation encoder.}
HistRISE uses the same point-cloud observation encoder as ChronoFlow Policy and only differs in the additional history track features.

\boldparagraph{History point tracks and track encoder.}
HistRISE takes normalized point tracks as history input, where $\texttt{num\_targets}=1$ and $\texttt{points\_per\_target}=5$ by default.
For fair comparison, we use the same track encoder design as ChronoFlow Policy for extracting history features: temporal patch embedding with $\texttt{patch\_size}=4$ and $\texttt{embed\_dim}=256$, followed by cross-attention pooling with a learnable query.
The cross-attention module uses $\texttt{query\_dim}=512$, $\texttt{num\_queries}=1$, $\texttt{n\_layers}=4$, $\texttt{n\_heads}=8$, $\texttt{ff\_dim}=1024$, dropout $0.1$, and time embedding.

\boldparagraph{Token fusion.}
The encoded track features are projected to the Transformer hidden dimension and appended to the point feature sequence before Transformer readout.
The readout feature conditions the same action diffusion head as RISE.
HistRISE is trained and evaluated with \textbf{action diffusion only}.

\subsection{Motion Before Action (MBA)}
\label{sec:impl-mba}

MBA~\cite{su2025motionactiondiffusingobject} uses a \emph{two-stage} diffusion pipeline in which an object-motion summary is first sampled and then used to condition action diffusion.
\textbf{Key differences to ChronoFlow Policy:} MBA does not model interaction-centric gripper-object keypoints.
Instead, it predicts object pose trajectories and does not use an explicit ChronoFlow history encoder.

\boldparagraph{Shared observation encoder.}
MBA uses the same observation encoder as ChronoFlow Policy to extract the current observation feature from the point cloud.

\boldparagraph{Two-stage diffusion.}
MBA first samples a future object-motion trajectory, stored as \texttt{obj\_pos} with dimension $d_o=6$, using a conditional Unet1D diffusion model.
It then generates actions using a second conditional Unet1D diffusion model conditioned on the predicted object motion.

\boldparagraph{Condition construction.}
For action generation, MBA embeds the predicted object-motion trajectory with an MLP of width \texttt{[32,32]} and concatenates it with the current observation feature, followed by a final projection to a 128-dimensional global condition.

\boldparagraph{Diffusion backbone.}
Both diffusion stages use \texttt{ConditionalUnet1D} with $\texttt{down\_dims}=(256,512,1024)$, diffusion embedding dimension $256$, kernel size $5$, group norm groups $8$, and FiLM conditioning by default.

\subsection{3D Flow Diffusion Policy (3D-FDP)}
\label{sec:impl-3dfdp}

3D-FDP~\cite{noh20253dflowdiffusionpolicy} is a two-stage diffusion policy that first predicts a future \emph{scene-level} point-cloud trajectory and then generates actions conditioned on the predicted future scene.
\textbf{Key differences to ChronoFlow Policy:} 3D-FDP predicts future dynamics at the entire-scene point-cloud level rather than using interaction-centric gripper-object keypoints, and it does not incorporate ChronoFlow history.

\boldparagraph{Shared observation encoder.}
3D-FDP uses the same observation encoder as ChronoFlow Policy to extract the current observation feature.

\boldparagraph{Stage I: future point-cloud diffusion.}
3D-FDP predicts a future point-cloud sequence over horizon $H$ with $N_{\textsc{pc}}$ scene points by diffusing flattened XYZ coordinates:
\begin{equation}
\bm{X}^{\textsc{fut}}_{t:t+H}
\in
\mathbb{R}^{H\times (N_{\textsc{pc}}\cdot 3)}.
\end{equation}
This stage conditions only on the current observation feature and does not use any history encoder.

\boldparagraph{Stage II: action diffusion conditioned on predicted scene feature.}
After sampling the future point cloud, 3D-FDP encodes it into a compact feature using a PointNet-style encoder with output dimension $\texttt{pred\_pc\_feature\_dim}=64$ and averages per-timestep features over the horizon.
The action diffusion model conditions on the concatenation of the observation feature and the predicted future-scene feature, with dimensions $128+64$ by default.

\boldparagraph{Diffusion backbone.}
Both point-cloud diffusion and action diffusion use \texttt{Unet1D} with $\texttt{down\_dims}=(256,512,1024)$, diffusion embedding dimension $256$, kernel size $5$, group norm groups $8$, and FiLM conditioning by default.
The point-cloud diffusion loss is weighted by $\lambda_{\textsc{pc}}=1$.

\section{Training Curves}
\label{sec:supp-training-curves}

To further analyze the learning behavior of different policies, we report training curves on representative tasks from MetaWorld and RoboTwin.
All methods follow the same evaluation protocol as in the main paper.
These curves complement the final success-rate comparisons by showing how quickly each method improves during training and how stable the learned policy is across evaluation checkpoints.

\begin{figure}[H]
\centering
\includegraphics[width=\linewidth]{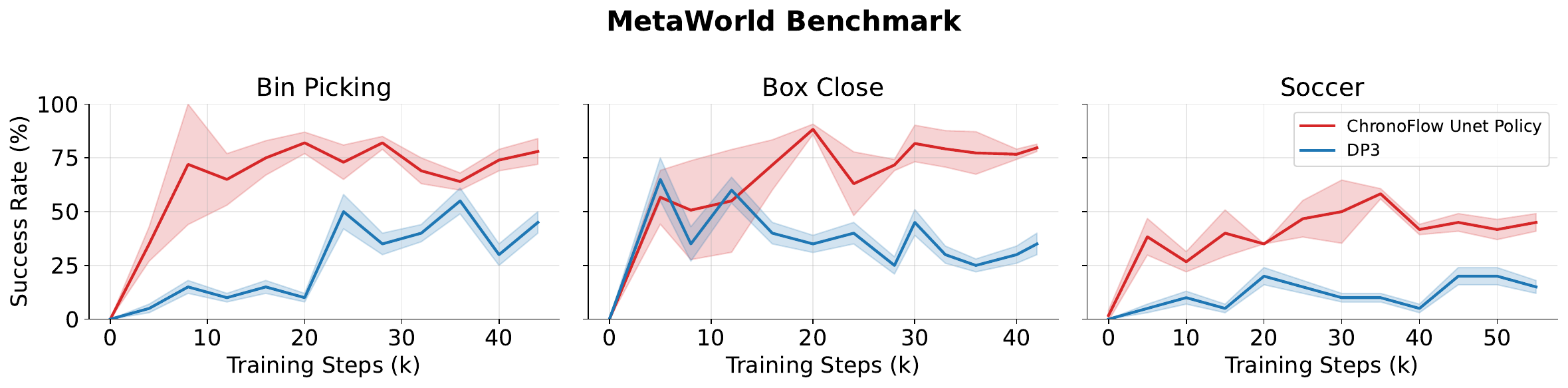}
\vspace{-0.6em}
\includegraphics[width=\linewidth]{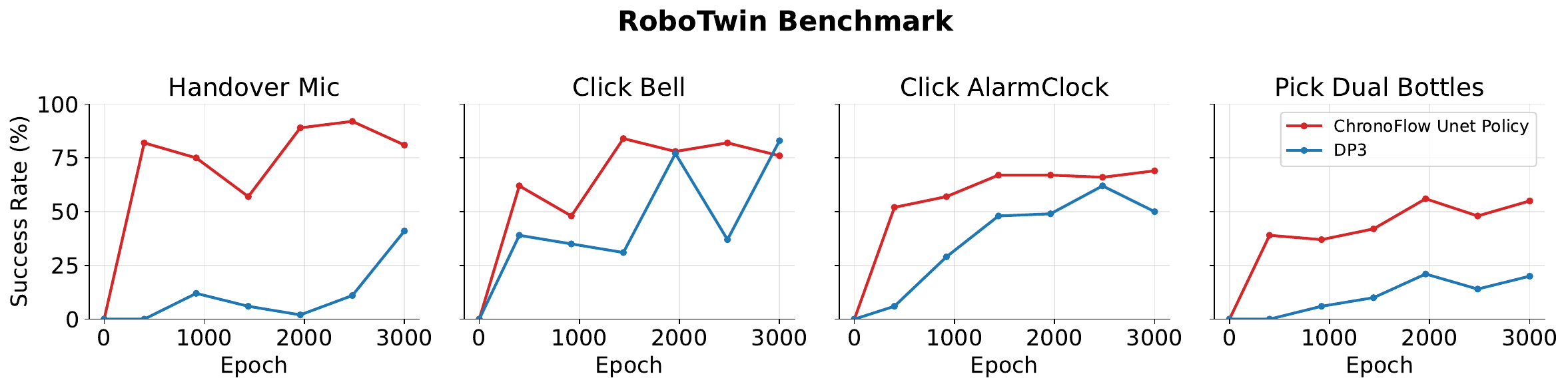}
\caption{
\textbf{Learning curves on simulation benchmarks.}
CFP (Unet) generally converges faster and achieves higher or comparable final success rates than DP3 across representative tasks, demonstrating the effectiveness of ChronoFlow supervision.
}
\label{fig:supp_learning_curves}
\end{figure}

As shown in Fig.~\ref{fig:supp_learning_curves}, CFP achieves faster convergence than DP3 on both single-arm and dual-arm manipulation tasks.
On MetaWorld, CFP rapidly improves on interaction-intensive tasks such as \textit{Bin Picking} and maintains a clear advantage throughout training.
On RoboTwin, CFP reaches high success rates earlier on long-horizon or coordination-heavy tasks such as \textit{Handover Mic} and \textit{Pick Dual Bottles}.
These results indicate that ChronoFlow provides a stronger training signal than action-only supervision, helping the policy learn object--gripper interaction dynamics more efficiently.

\end{document}